\documentclass[journal]{IEEEtran}
\usepackage{graphicx}
\usepackage[export]{adjustbox}
\usepackage{amsmath,amssymb,amsfonts}
\usepackage{hyperref}

\ifCLASSINFOpdf
\else
\fi
\begin{document}

%
\title{Towards Model Predictive Control for Acrobatic Quadrotor Flights}
%
%
%

\author{Saransh Jain,
        Yash Shethwala,{}
        and~Jnaneshwar Das{}
\thanks{S. Jain, Y. Shethwala, and J. Das was with the Arizona State University, AZ, 85281 USA e-mail: (see sjain164@asu.edu).}
\thanks{}
\thanks{}}

%
%

\markboth{Journal of \LaTeX\ Nov~2023}%
{Shell \MakeLowercase{\textit{et al.}}: Bare Demo of IEEEtran.cls for IEEE Journals}
%



\maketitle

\begin{abstract}

This study explores modeling and control for quadrotor acrobatics, with a focus on execution of flip (or belly-flop) maneuvers. Flip maneuvers are an elegant way to execute sensor probe delivery into no-fly or hazardous zones such as volcanic vents. Successfully navigating flips necessitates dynamically feasible trajectories and precise control, factors influenced by rotor dynamics, thrust allocation, and strategic control methodologies. To address these challenges, the research introduces a novel approach that utilizes Model Predictive Control (MPC) for real-time trajectory planning and execution. The MPC formulation takes into account dynamic constraints and environmental variables, ensuring system stability during aggressive maneuvers.
The effectiveness of the proposed methodology is rigorously examined through simulation studiesin ROS and Gazebo environments, providing valuable insights into quadrotor behavior, response time, and trajectory tracking accuracy. Real-time flight experiments on a custom agile quadrotor drone using flight controllers like PixHawk 4 and Hardkernel Odroid companion computer validate the practical applicability of the MPC-designed controllers. The experiments not only confirm the successful execution of the proposed approach but also demonstrate its adaptability to dynamic real-world scenarios.
The outcomes of this research contribute significantly to the advancement of autonomous aerial robotics, specifically aerial acrobatics for enhanced mission capabilities. Beyond the application of MPC controllers for precise and autonomous probe throws into no-fly zones like volcanic vents, practical implications include optimal image capture views of points of interest through efficient flight paths, for example, full roll maneuvers during flight. This research paves the way for further exploration and utilization of quadrotors in demanding and dynamic scenarios, showcasing the potential for groundbreaking applications in various fields.
Video Link: \url{ https://www.youtube.com/watch?v=UzR0PWjy9W4}

\end{abstract}


%
\IEEEpeerreviewmaketitle

\section{Introduction}
%
%
%
%
Unpiloted Aircraft Systems (UAS) have revolutionized global industries, reshaping the landscape of various tasks and operations that are hazardous or inaccessible for humans. These versatile flying platforms, especially the quadrotor format, have proven invaluable in applications spanning environmental monitoring, infrastructure inspection, and logistics. Quadrotors have a distinct four-rotor configuration, and posses exceptional maneuverability, stability, and adaptability in navigating complex environments. Hence autonomous quadrotors can be deployed in life critical scenarios, such as searching for survivors in damaged buildings, entering and clearing buildings during emergency situations, and collecting information in hazardous biological or nuclear contaminated sites. Creating situational awareness without putting humans in danger is extremely valuable. 
\begin{figure}[!ht]
{\includegraphics[width =8.5 cm ]{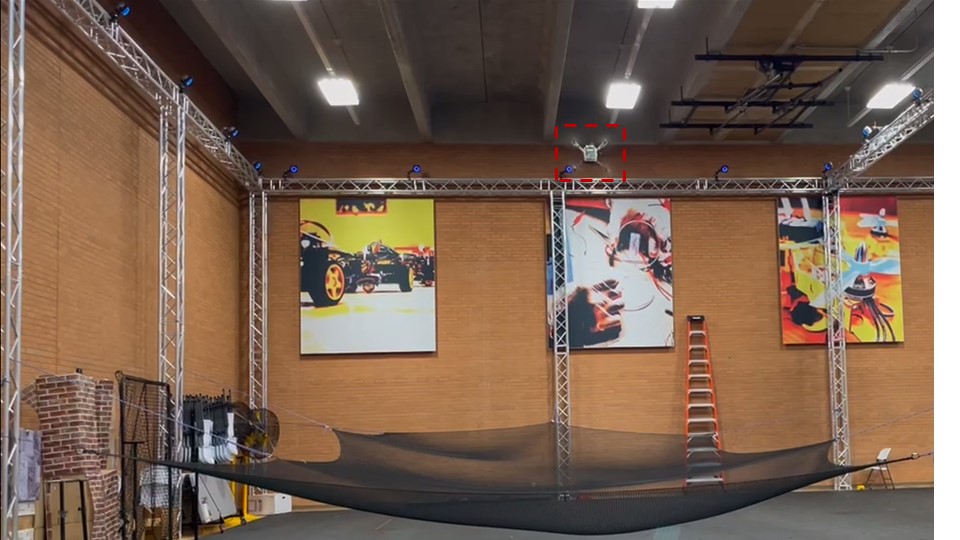}}
\caption{Quadrotor performing flip maneuver}
\label{fig}
\end{figure}
Extensive studies have investigated aggressive maneuvers and their applications. For instance, tight turns, achieved by swiftly adjusting the yaw, pitch, and roll angles, enable quadrotors to navigate through narrow spaces or execute sharp changes in direction. Mellinger et al. (\cite{perching}) designed dynamically feasible trajectories and control strategies that enable quadrotors to navigate through vertical and horizontal openings while achieving desired orientations. Other researchers, such as (\cite{abbeel2008apprenticeship}), (\cite{gillula2010design}) Gillula et al. (2010), and (\cite{lupashin2010simple}) Lupashin et al. (2010), have demonstrated aggressive aerial maneuvers with small-scale rotorcraft.

Considerable efforts in this field focus on developing strategies for generating sequences of controllers that stabilize the quadrotor to a desired state, capitalizing on its ability to change direction and attitude swiftly. For instance, (\cite{l2018introduction}) showcases high-speed navigation with dynamic obstacle avoidance, demanding the generation and control of significant linear and angular accelerations, among other features of aggressive maneuvering. These findings underscore the need for high-performance controllers that can effectively handle uncertainties, disturbances, unmodeled dynamics, and the coupled effects of quadrotor dynamics (\cite{madani2006backstepping}).
\par
In the pursuit of designing high-performance control algorithms, (\cite{goodarzi2013geometric}) Goodarzi et al. developed a nonlinear PID controller for quadrotors in SE 3 (six-dimensional space). Their research aimed to analyze the controller's performance during complex maneuvers. The paper by Goodarzi et al. (\cite{goodarzi2013geometric}) presents a novel approach to control algorithm design in the context of quadrotors. However, standalone PID controllers and feedback controllers such as LQR cannot handle the system's nonlinearities and uncertainties.

In a related study, the authors of the (\cite{perching}) formulated a high-gain control strategy for trajectory tracking and generation during aggressive maneuvers. This research focused on enhancing the maneuverability and control precision of quadrotors.

Another significant contribution to the field was made by Lee et al. (\cite{flores2013lyapunov}), who designed a robust control system tailored explicitly for quadrotors. Their approach aimed to address the challenges posed by uncertainties and disturbances in the system.

Furthermore, an adaptive controller was proposed by another research team (\cite{lei2018robust}). The adaptive controller aimed to improve the performance and stability of quadrotors by adapting to changing conditions and varying flight dynamics.

Despite these controllers' promising performance and global stability claims, they exhibit limitations in handling system and task-specific constraints essential for real-time drone testing. In the probe throw through quadrotor flip maneuvers, it is necessary to assign limitations on the release angle of the investigation to attain desired range. Also, we needed to consider system limitations as constraints in controller design. The controllers discussed above cannot view these constraints, which can significantly impact the drone's performance and safety during flight.

Moreover, the controllers discussed above require prior information about the complete trajectory, which hampers their versatility. In scenarios where uncertainty and discontinuity exist in the designed course, the performance of these controllers can be adversely affected.

In addition to physics-based controllers, researchers have also explored the implementation of data-driven controllers for performing aggressive maneuvers. For instance, (\cite{bisheban2020geometric}) designed a neural network-based controller to train the system for aggressive maneuvers. Similarly, (\cite{hwangbo2017control}) proposed a reinforcement learning-based approach for trajectory tracking of quadrotors. These controllers have demonstrated promising results in terms of maneuverability and performance.

However, it is essential to note that ensuring the system's stability is challenging when employing data-driven controllers. Unlike physics-based controllers, typically designed with a solid theoretical foundation, data-driven controllers rely on learned patterns from training data. Consequently, the system's stability may be more difficult to guarantee with data-driven approaches. 

The trajectory tracking of quadrotors has been extensively investigated by numerous researchers, as evidenced by studies such as \cite{trajec}, \cite{traj2}, and \cite{traj3}. Additionally, aggressive trajectory tracking has garnered considerable attention, as highlighted in research works like \ quote {perching}, \cite{chen2019controller}, and \cite{gomaa2022computationally}. While these studies have contributed valuable insights to the field, performing multiple iterations of a complex maneuver, such as a flip, to demonstrate the repeatability and reliability of the controller remains highly challenging. This work answers the challeges such as: 

Quadrotor dynamics with probe: Deriving combined dynamics is challenging due to the underactuated nature of the system, leading to complex nonlinear dynamics that complicate control.

Control design: Designing a controller for the nonlinear dynamics of a quadrotor with varying mass during flips poses additional challenges, requiring innovative solutions for effective control design and closed-loop stability.

Robustness of MPC: Ensuring the repeatability and reliability of the Model Predictive Controller (MPC) in real-time, while handling uncertainties and disturbances, becomes a formidable challenge, adding complexity to the task.

Flip and throw maneuver: Despite nonlinear dynamics challenges, the flip and throw maneuver introduces further complexity due to varying mass. Designing a controller capable of adapting to changing dynamics and ensuring stability becomes intricate, necessitating sophisticated control strategies.

This research paper proposes the adoption of Model Predictive Control (MPC) algorithms for the execution of acrobatic maneuvers, specifically focusing on quadrotor flips and throws. The controller's performance and robustness were rigorously evaluated through both simulation and real-time flight experiments.

To assess the repeatability and reliability of the proposed control design, a series of ten real-time flight trials were conducted, specifically focusing on the quadrotor flip maneuver. The integration of both simulation and real-world flight data enhances the credibility and applicability of the findings, providing valuable insights for the implementation of MPC for dynamic and complex aerial maneuvers.

\section{Mathematical Modeling}
The dynamics of the quadrotors with a cable-suspended probe is derived using the Lagrangian method. Further, this model is used for the controller design. In this model, the quadrotor is treated as a single-point mass capable of moving within inertial coordinates. On the other hand, the suspended probe is assumed to be attached to the quadrotor's center of gravity. To simplify the analysis, the link connecting the quadrotor and the payload is assumed to be massless and rigid, with a fixed size that cannot be changed. The payload is considered as a point mass. 

According to Fig. 2, the following coordinates frames are considered. Origin of inertial coordinate frame $O_g$ = $[x_g, y_g, z_g]$, origin of the body-fixed frame is given by $O_b$ = $[x_b, y_b, z_b]$ which in fact the center of mass coordinates of the quadrotor. The coordinate frame of the payload is $O_b$ = $[x_b, y_b, z_b]$. The position coordinate of the quadrotor with respect to the inertial frame is $P_q = [x,y,z]^T$. Rotation: The quadrotor's roll, pitch, and yaw motion is represented by $Q_q= [\theta, \phi, \psi]$, respectively. Further, the rotation of the link between the quadrotor and payload can swing in the plane and perpendicular to the plane denoted by: $[\alpha, \beta]$, respectively. $l$ is the length of the link. 

\begin{figure}[htbp]
{\includegraphics[width =9 cm ]{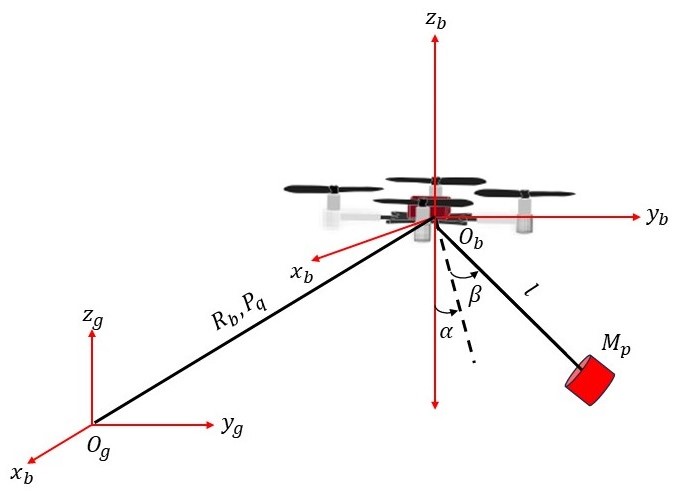}}
\caption{Example of a figure caption.}
\label{fig}
\end{figure}
The rotational matrix from the body frame of the quadrotor to the inertial frame is given by:
\begin{align}
\mathbf{R_b} = Rt(z,\psi)Rt(y,\phi)Rt(x,\theta) \nonumber
\end{align}
Where $Rt(z,\psi), Rt(y,\phi)Rt(x,\theta)$ is the rotation matrices about $z$, $y$ and $x$ axis, respectively. 
Let $\mathbf{P_q}$ and $\mathbf{P_l}$ be the position vector of the quadrotor and the position vector of the link with respect to the inertial frame. 
\begin{align}\label{Eq2}
    \mathbf{P_q}= \mathbf{R_b}\mathbf{^bP_q}
\end{align}
where $\mathbf{^bP_q}$ is the quadrotor position in the body frame.
\begin{align}\label{eq.3}
    \mathbf{P_l} = \mathbf{P_q} + \mathbf{R_b}\mathbf{^bP_l}
\end{align}
To derive the Lagrangian, an expression of the kinetic and potential energy can be computed as:
\begin{align}\label{eq.5}
    K = \frac{1}{2}M_q\Dot{P_q}^T\Dot{P_q} + \frac{1}{2}{\Omega_q^T}R_bIR_b^T\Omega_q + \frac{1}{2}M_p\Dot{P_l}^T\Dot{P_l} \\ \nonumber+ \frac{1}{2}M_pl^2(\dot{\alpha}^2+\dot{\beta}^2)
\end{align}
where $M_p$ and $M_p$ are the mass of the quadrotor and probe, respectively. $\Omega_q$ is the angular velocity of quadrotor in body frame, $I$ is the moment of inertia of quadrotor.
Now, the potential energy comes out to be:
\begin{align}\label{eq.6}
    V = M_qgz + M_pg(z-lcos(\alpha))
\end{align}
By applying the Euler-Lagrange formulation,
\begin{align}\label{Eq.7}
\frac{d}{dt}(\frac{\partial L}{\partial\mathbf{\Dot{q}}})-(\frac{\partial L}{\partial\mathbf{{q}}})=\boldsymbol{{\tau}}
\end{align}
where \textbf{q}, $\dot{\mathbf{q}}$, and $\boldsymbol{{\tau}}$ are the generalized position, velocity, and  the control input of the system, respectively, with:

\begin{align}\label{Eq.8}
\mathbf{q}=\begin{bmatrix}
{x}\\{y}\\{z}\\{\theta}\\{\phi}\\{\psi}\\{\alpha}\\{\beta}  \end{bmatrix}
\end{align}
After solving equation (\ref{Eq.7}). The equation of motion of the integrated quadrotor and payload system comes out to be:
\begin{align}\label{Eq.10}
    \Dot{P_q} = V_q
\end{align}
where $V_q$ is the linear velocity of the quadrotor
\begin{align}\label{Eq.11}
    (M_q+M_p)(\Dot{V_q}+ge_3) = (^bP_lfR_be_3 - M_ql(\Dot{^bP_l}\Dot{^bP_l}))^bP_l
\end{align}
\begin{align}\label{Eq.12}
    \Dot{^bP_l} = (\Omega_p) \times (^bP_l)
\end{align}
\begin{align}\label{Eq.13}
    M_ql\Dot{\Omega_p} = -(\Dot{^bP_l}) \times fR_be_3
\end{align}
\begin{align}\label{Eq.14}
    \Dot{R_b} = R_b\hat{\Omega_q}
\end{align}
\begin{align}\label{Eq.15}
    I\Dot{\Omega_q} + \Omega_q \times I\Omega_q = M
\end{align}
where $M$ is the moment.
\section{Control Design}
In this work, Model Predictive Control (MPC) is used to design the controller for optimal trajectory generation and tracking. MPC controller predicts future behavior for a specified finite time horizon using the system's dynamic model. This prediction is utilized to determine the optimal control strategy for the given time horizon that minimizes the cost function subjected to given constraints. The calculated control inputs are then applied to the plant model, enabling new system measurements to be obtained. This iterative process continues recursively, with continuously updated predictions and control actions based on the evolving system dynamics. 
The MPC controller formualtion for the quadrotor flip manuevers.
\begin{align}
    \mathbf{x} = [P_q, \Dot{P_q}, \Ddot{P_q}]^T
\end{align}
\begin{align}
    \mathbf{u} =f 
\end{align} 

Now, the cost function of the MPC controller can be designed as follows:
\begin{align}\label{MPC_law1}
\underset{U}{\text {min}}\hspace{5pt}\mathbf{J_0(x_0,U,x_{ref})} &= \underset{k=0}{\sum^{N-1}} (\mathbf{x_k -x_{ref,k}^T})\mathbf{Q}(\mathbf{x_k -x_{ref,k}}) \nonumber \\ + (\mathbf{u_k -u_{k-1}^T})R(\mathbf{u_k -u_{k-1}}) \nonumber&+ (\mathbf{x_N -x_{ref,N}^T})\mathbf{P}(\mathbf{x_N -x_{ref,N}})& \\ \nonumber
\text{subject to }  &\Dot{x} =  f(x(t), u(t)) \\& \nonumber
u(t)\in U \\& 
x(t) \in X& 
\end{align}
where the control input constraints define as:
\begin{align}
    \{ U \in R^m \mid (u_{min})\leq u(k) \leq (u_{max}) \}
\end{align}
The value of $N$ =10 and the time step $\delta$ = 0.04. The $Q$, $P$ and $R$ matrix is defined as
 \begin{align}
    \mathbf{Q} = diag[50,10,20,10,20,10,10,10,10]
 \end{align}
 \begin{align}
     \mathbf{P} =diag[100,100,10,10,10,10,10,10,10]
 \end{align}
 \begin{align}
     R =1
 \end{align}
The attitude and attitude rate command can be computed using the desired direction, which is the third body-fixed frame $\Vec{{b_3}_d}$. Assuming that $\Vec{{b_1}_d}$ is not parallel to $\Vec{{b_3}_d}$ and the projection onto the plane normal to third body-fixed $\Vec{{b_3}_d}$ frame gives the complete desired attitude: ${{R_b}_d} =[\Vec{{b_2}_d}\times \Vec{{b_3}_d}, \Vec{{b_2}_d},\Vec{{b_3}_d}] \in SO(3)$ (\cite{geometric}). 
 
 $\Vec{{b_3}_d}$ is defined as:
 \begin{align}
     \Vec{{b_3}_d} = \frac{u-M_qge_3}{\parallel{u-M_qge_3}\parallel}
 \end{align}
 \begin{align}
     {R_b}_d= [\Vec{{b_2}_d}\times \Vec{{b_3}_d}, \Vec{{b_2}_d},\Vec{{b_3}_d}]
 \end{align}
 \begin{figure}[h!]
\centering
\includegraphics[width =8.5 cm ]{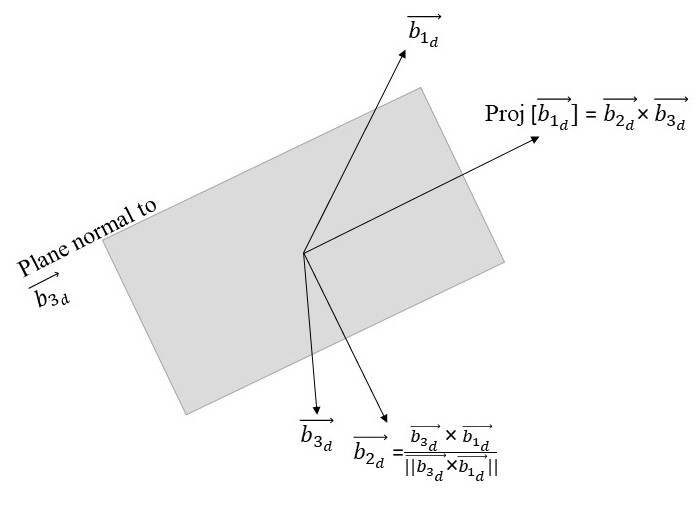}
\centering
\caption{Desired Heading Direction \cite{geometric}.}
\label{figure:heading}
\end{figure}
 The attitude rate can be calculated as follows:
 \begin{align}
     {e_R}_b = \frac{1}{2}({R_b}_d^TR_b -R_b^T{R_b}_d)^\vee
 \end{align}
 \begin{align}\label{4.17}
     \Omega_q = \frac{2}{\tau_\Omega}{e_R}_b
 \end{align}
 
 where $\tau_\Omega$ is the first-order system time constant, the commanded acceleration from the position loop $f$ and the commanded yaw $\psi_d$ angle have to be converted into a desired attitude The equation (\ref{4.17}) is adopted from \cite{ETHattitude}
\section{Flip and Throw Maneuver}
The flip and throw maneuver comprises three distinct stages, each playing a crucial role in achieving a successful outcome. In the first stage, the quadrotor, carrying the probe, diligently navigates its way to the predetermined rally point. This initial stage is essential for positioning the quadrotor and preparing it for subsequent maneuvers.

Moving into the second stage, the quadrotor executes the flip maneuver. The third stage is probe release and the recovery of the quadrotor. The flip is a complex aerial maneuver that requires precise control and coordination of the quadrotor's movements. Understanding the dynamics and intricacies of the flip is paramount as it sets the stage for the subsequent throw.

The recovery process following the flip is particularly critical in the flip and throw maneuver. They are stabilizing the quadrotor after the flip becomes significantly more challenging due to the inherent complexities introduced by the throw. The dynamics of the quadrotor during the flip are influenced by factors such as varying mass, which directly impact the control and stability of the aircraft. Furthermore, throwing the probe occurs a recoil or the reaction force on the quadrotor, further exacerbating the difficulties of recovery.

Now, the MPC (Model Predictive Control) controller comes into play, generating an adequate pitch rate to propel the probe to a predetermined distance while ensuring a safe recovery following the flip. The controller's role is to achieve the desired throw and to counterbalance the effects of the varying mass and attenuated reaction force during the recovery phase. This requires meticulous planning and precise control algorithms to achieve the desired outcome.
\begin{figure}[h!]
\centering
\includegraphics[width =9.4cm ]{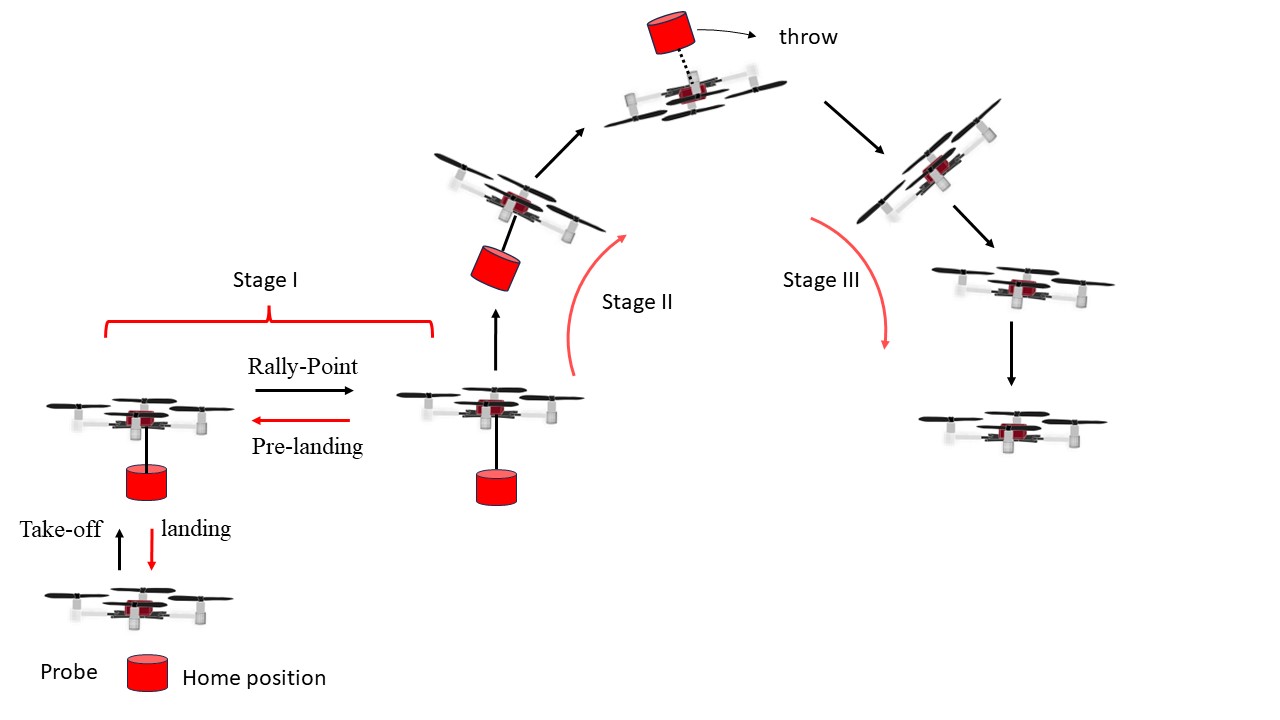}
\centering
\caption{Stages and Scenario of Flip-Throw Maneuver Mission}
\label{flip_throw}
\end{figure}
The three stages of the flip and throw maneuvers are described in Fig. 
\ref{flip_throw}.

\begin{figure}[h]
\centering
\includegraphics[width =8 cm ]{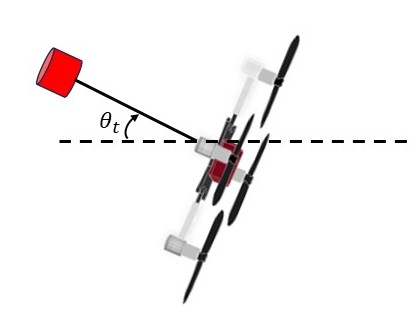}
\centering
\caption{Throw Angle of the Probe}
\label{angle}
\end{figure}
In fig. \ref{angle}, $\theta_t$ denotes the probe throw angle. During the MPC optimization, the probe angle and the speed are also calculated based on the desired range. 
\begin{align}\label{eq.range}
    R = \frac{Vcos(\theta_t)(Vsin(\theta_t) + \sqrt{Vsin(\theta_t)^2 +2gh})}{g}
\end{align}
In Equation \ref{eq.range}, $R$ represents the desired range, $V$ denotes the velocity of the probe, and $h$ represents the height of the probe from the ground. Constraints were imposed on both $V$ and $\theta_t$ during the optimization. The pitch rate of the quadrotor directly influences the velocity $V$. Considering the desired range and the height limitations, the optimal velocity $V$ and throw angle $\theta_t$ can be computed to achieve the desired distance. Imposing constraints on the velocity $V$ is necessary due to observations made during the flip maneuver, where a lower pitch rate results in longer recovery times or even potential crashes.

\section{Simulation}
The performance of the proposed controller was first tested in the simulation environment. A Robotic Operating System (ROS) node was created and the simulation was performed in the Gazebo environment. In order to solve the optimization problem, the CasADi toolbox is employed. This toolbox is known for its high efficiency in implementing optimization problems and compatibility with Python 3. In the CasADi toolbox, Sequential Quadratic Programming (SQP) method is implemented to solve the optimization because SQP can handle the nonlinear constraints. It can effectively handle both equality and inequality constraints. SQP methods are known for their robustness, especially in the presence of noise or uncertainties in the problem formulation. They can handle ill-conditioned problems and adjust the step sizes in each iteration, ensuring stability and reliable convergence even in challenging scenarios \cite{SQp}. 

To address the issue of the feasibility of the MPC solver, if the optimization solver fails to solve the optimization problem or cannot find the optimal solution at a given time step, the control strategy from the previous feasible optimization is employed to ensure the continuity of the process.

A mission scenario is designed as shown in Fig. \ref{flip_throw} for a quadrotor to evaluate the performance of the Model Predictive Control (MPC) controller. The mission involves a sequence of stages, starting with the quadrotor taking off from its initial home position and reaching a specified altitude. Subsequently, the quadrotor proceeds toward a designated rally point, where it initiates a series of flip maneuvers.
At the rally point, the quadrotor executes three distinct stages of the flip maneuver: the ascending phase, the flip and throw phase, and the recovery phase. These stages allow the quadrotor to perform a complete flip and throw the probe while maintaining control and stability. Upon successfully completing the flip and throw, the quadrotor stabilizes itself in an upside-down position at the rally point, holding its position.
After achieving stability, the quadrotor then navigates back to its initial home position, following a predefined trajectory, thus concluding the current mission. This process sets the stage for subsequent tasks to be initiated from the home position, ensuring a continuous cycle of mission execution.

The reference state of the mission stages is considered as follows: 
Take-off reference where $q$ is the orientation quatrains
\begin{align}
    x_{ref} = [0,0,2,0,0,0,0,0,0]^T, q_d  =[0,0,0,1], {\Omega_q}_d = [0,0,0]  
\end{align}

Rally-point reference
\begin{align}
    x_{ref} = [4,0,2,0,0,0,0,0,0]^T, q_d  =[0,0,0,1], {\Omega_q}_d = [0,0,0] 
\end{align}
Flip reference
\begin{align}
{\Omega_q}_d = [0,5\pi,0] rad/s
\end{align}
Hold position reference
\begin{align}
    x_{ref} = [4,0,2,0,0,0,0,0,0]^T, q_d  =[0,0,0,1], {\Omega_q}_d = [0,0,0] 
\end{align}

\begin{figure}[h]\label{controlstr}
\centering
\includegraphics[width =8.5 cm ]{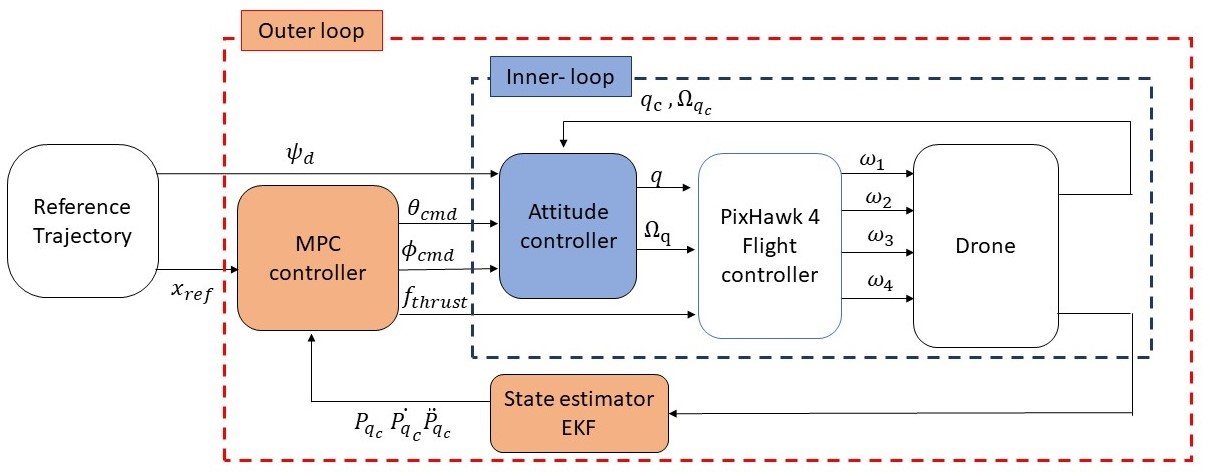}
\centering 
\caption{ Inner and Outer loop of Control Design of the Quadrotor}
\end{figure}
The control design for the quadrotor is shown Fig. 5.  proposed controller was evaded in the ROS and Gazebo environment, and the simulation results of the MPC controller on the mission scenario are shown in Fig. \ref{flip_throw}.

\begin{figure}[h!]
\centering
\includegraphics[width =8.5 cm ]{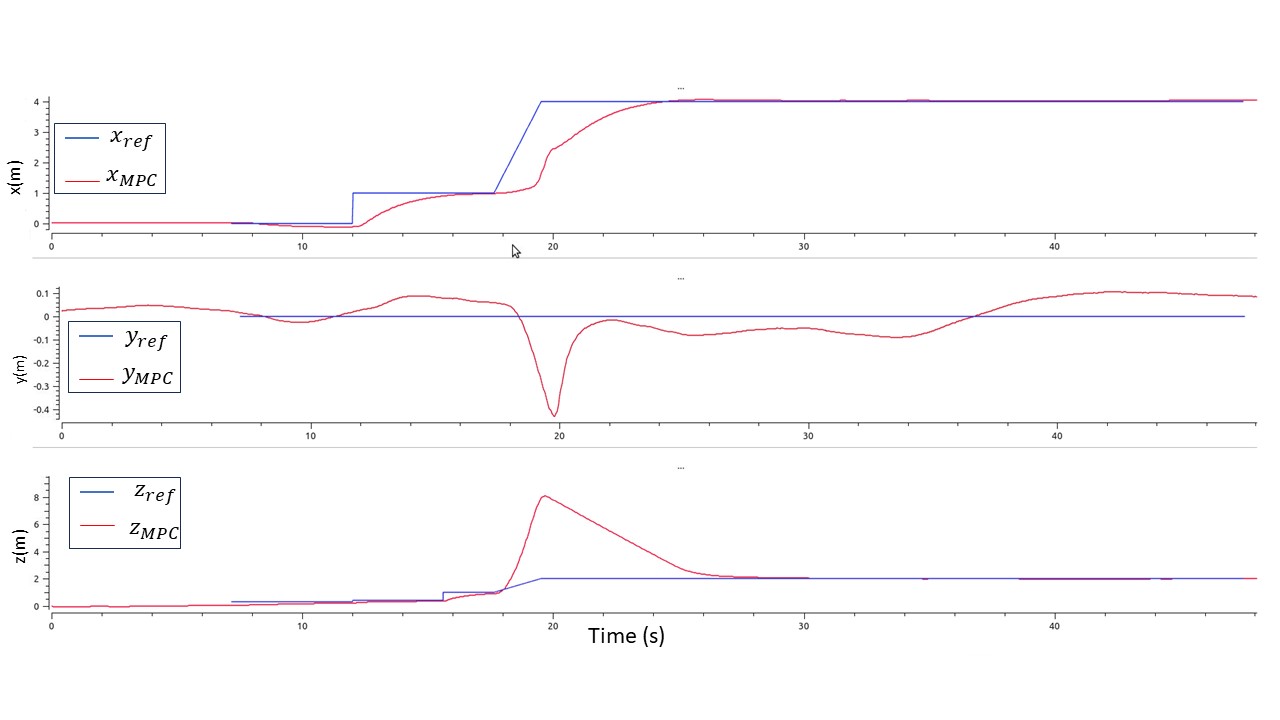}
\centering
\caption{Quadrotor Pose X, Y, and Z Response with MPC controller}
\label{probepose}
\end{figure}

\begin{figure}[h!]
\centering
\includegraphics[width =8.5 cm ]{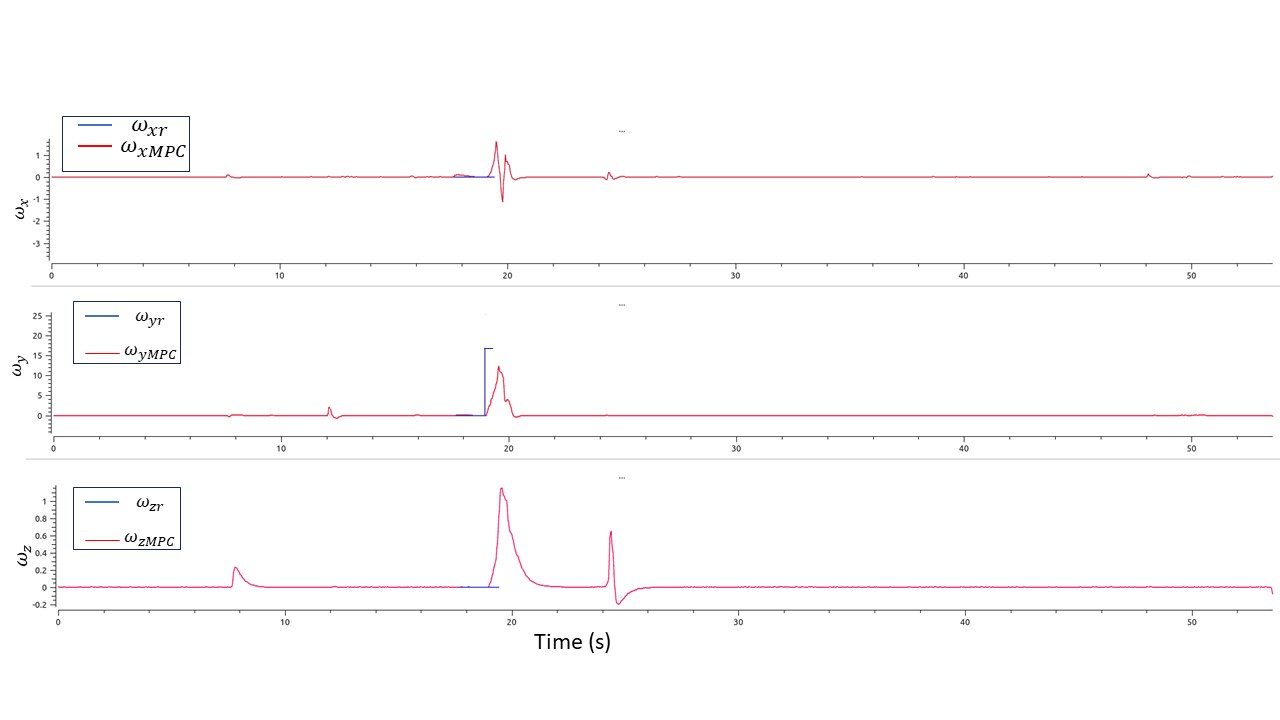}
\centering
\caption{Quadrotor Attitude Rate Response with MPC controller}
\label{probethrow}
\end{figure}

\begin{figure}[h!]
\centering
\includegraphics[width =8.5 cm ]{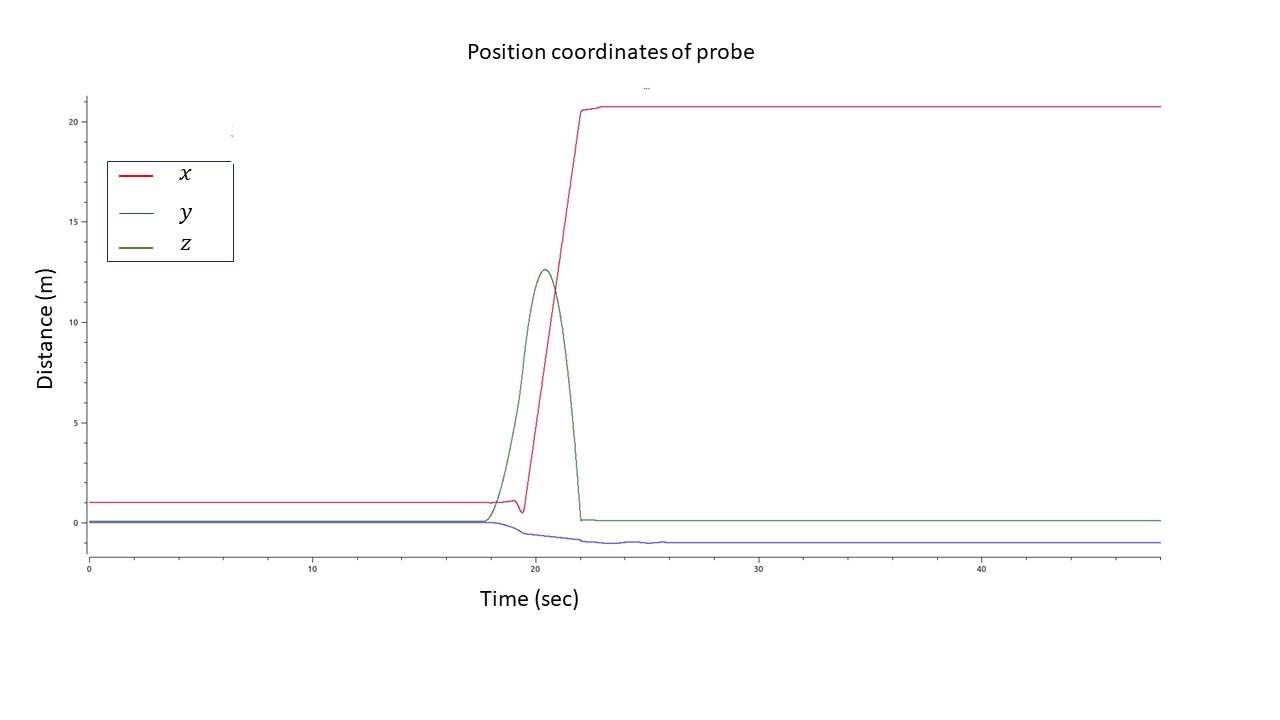}
\centering
\caption{Probe Position Response in X, Y, and Z Direction}
\label{range}
\end{figure}
The parameters of the quadrotor remained unchanged for the flip maneuver, with a probe mass of $M_q = 0.2 kg$ and a link length of $l = 0.5 m$. The performance of the MPC controller for the probe throw maneuver is presented in Fig. \ref{probepose} and Fig. \ref{probethrow}. The MPC controller successfully executed the mission, as evidenced by the results.

During the flip around the 15-second mark, the quadrotor successfully threw the probe and managed to recover to the hold position coordinate, i.e., $(4, 0, 2)$. The desired throw range was set to 20 m from the probe's initial coordinates at $(1, 0, 0)$. The probe's trajectory from its initial position to the final landing coordinates is depicted in Fig. \ref{range}.

The desired throw range of 20 m was specifically targeted in the x-direction. As shown in Fig. \ref{range}, the final x-position of the probe is 21 m, indicating that the probe reached the desired position accurately.

This successful mission performance provides further evidence of the effectiveness of the proposed control scheme for precise tracking, safe execution of the flip maneuver, and accurate probe throw at the desired location.

\section{Experiments}

The performance of the Model Predictive Control (MPC) controller is comprehensively evaluated through a series of real-time experiments. The objective is to assess the efficacy and robustness of the MPC controller in practical scenarios. To facilitate this evaluation, a detailed description of the hardware setup employed for the experiments is provided, ensuring transparency and repeatability of the research.

In the actual experimental trials, only the flip maneuver was assessed, primarily due to logistical limitations associated with executing the door flip in an outdoor setting and concerns regarding potential damage to the Motion Capture System cameras resulting from the probe throw.
\begin{figure}[h!]
\centering
\includegraphics[width =8.5 cm ]{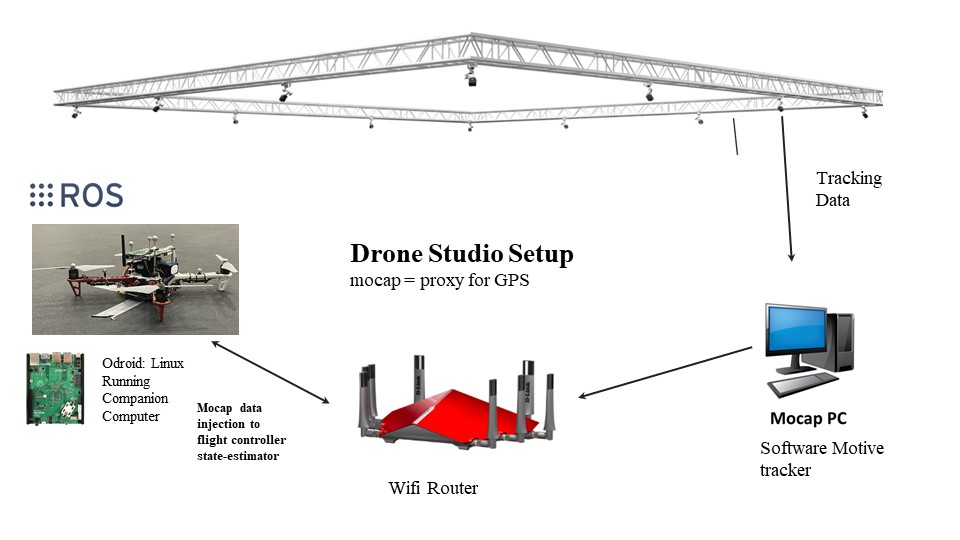}
\centering
\caption{Hardware Setup}
\label{figure:hardware}
\end{figure}

\begin{figure}[h!]
\centering
\includegraphics[width =8.5 cm ]{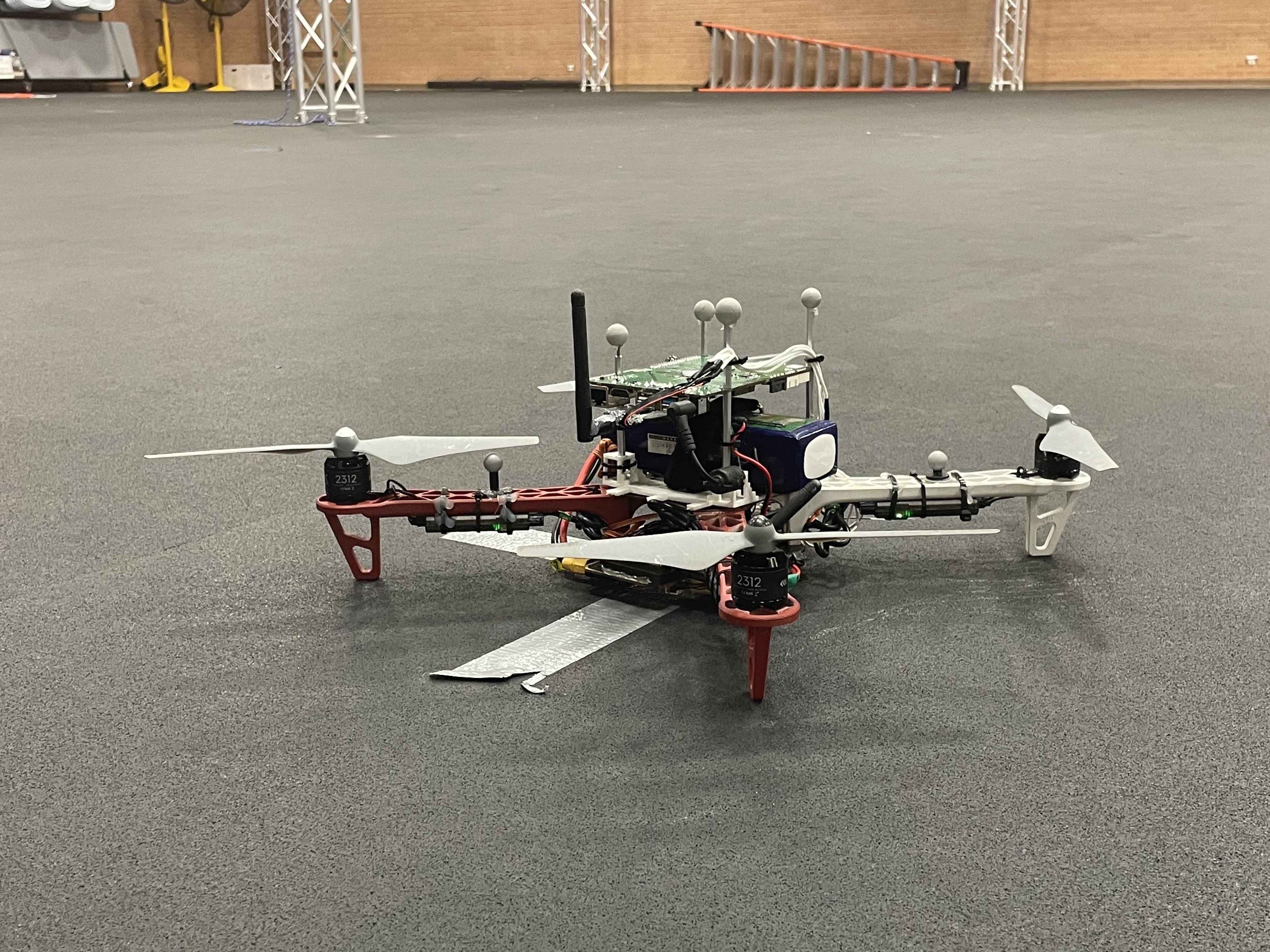}
\centering
\caption{Quadrotor: Flippy}
\label{figure:drone}
\end{figure}

 The experiments were performed in the ASU drone studio, with 114 high-precision cameras and a 3D tracking system. A motion capture system is utilized as a proxy to simulate the positioning system typically reliant on GPS.  Real-time autonomous flight conducted using the lightweight custom quadrotor equipped with PixHawk 4 flight controller, ROS on a companion computer: Hardkernel Odroid M1. The motion capture and the companion computer are connected by the established local network running at 5 Ghz band.
The experiment setup's outline is shown in Fig.\ref{figure:hardware}.

To assess the robustness of the MPC controller, a series of eight flip maneuvers are performed and thoroughly tested. These flip maneuvers are carefully designed to evaluate the controller's ability to handle rapid and dynamic changes in flight orientation. The MPC controller's performance during each trial of the flip maneuver is closely monitored and recorded.

The performance data obtained from the flip maneuvers are plotted and analyzed in detail. The analysis allows for a comprehensive understanding of the controller's strengths, limitations, and areas for improvement. Additionally, the performance data analysis provides quantitative results that can be compared against the simulation results obtained in Gazebo. This comparison aids in verifying the consistency and reliability of the simulation model, further enhancing the confidence in the real-world applicability of the MPC controller.

The detailed analysis of the MPC controller's performance in each trial of the flip maneuvers contributes to a comprehensive evaluation of its robustness and suitability for practical applications. The insights gained from this analysis serve as a foundation for performing the flip and probe throw maneuver.

\begin{figure}[h!]
\centering
\includegraphics[width =8.5 cm ]{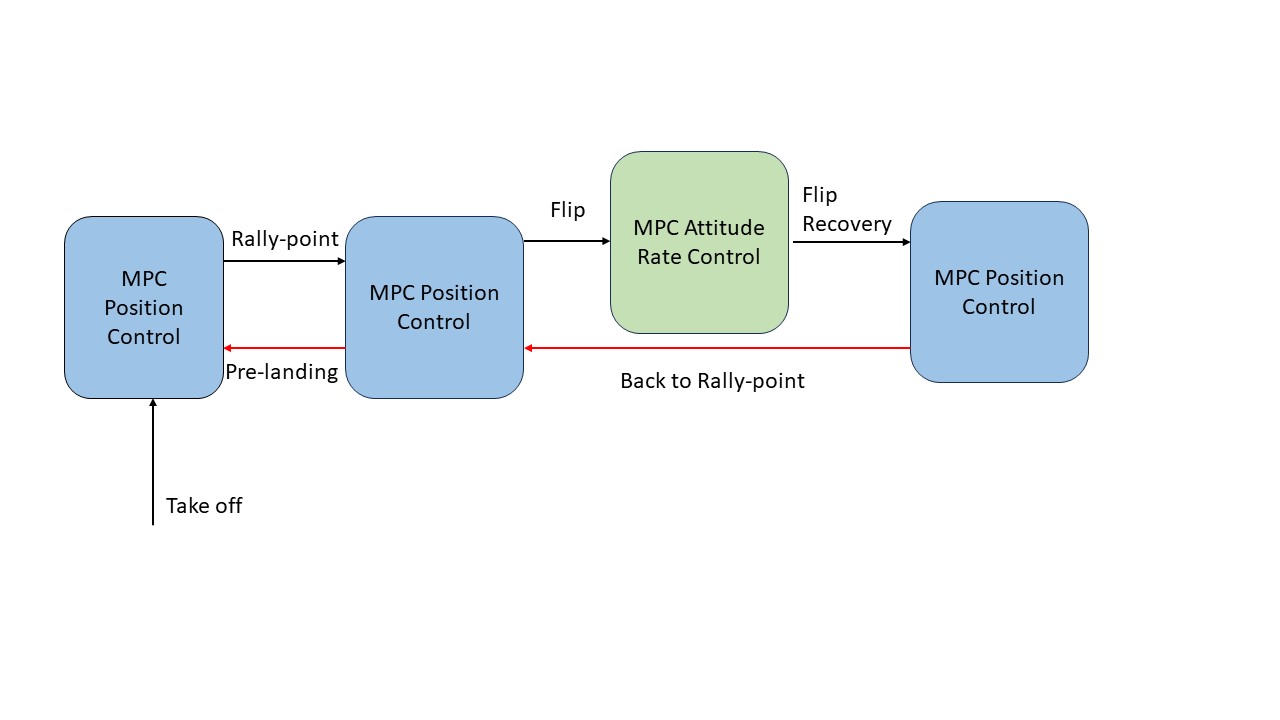}
\centering
\caption{Control strategy of the Flipping Maneuver Mission}
\label{cotrolstare_figure}
\end{figure}

The control strategy implemented in the experiments conducted is shown in Fig. \ref{cotrolstare_figure}.

\begin{figure}[h!]
\centering
\includegraphics[width =8.5 cm ]{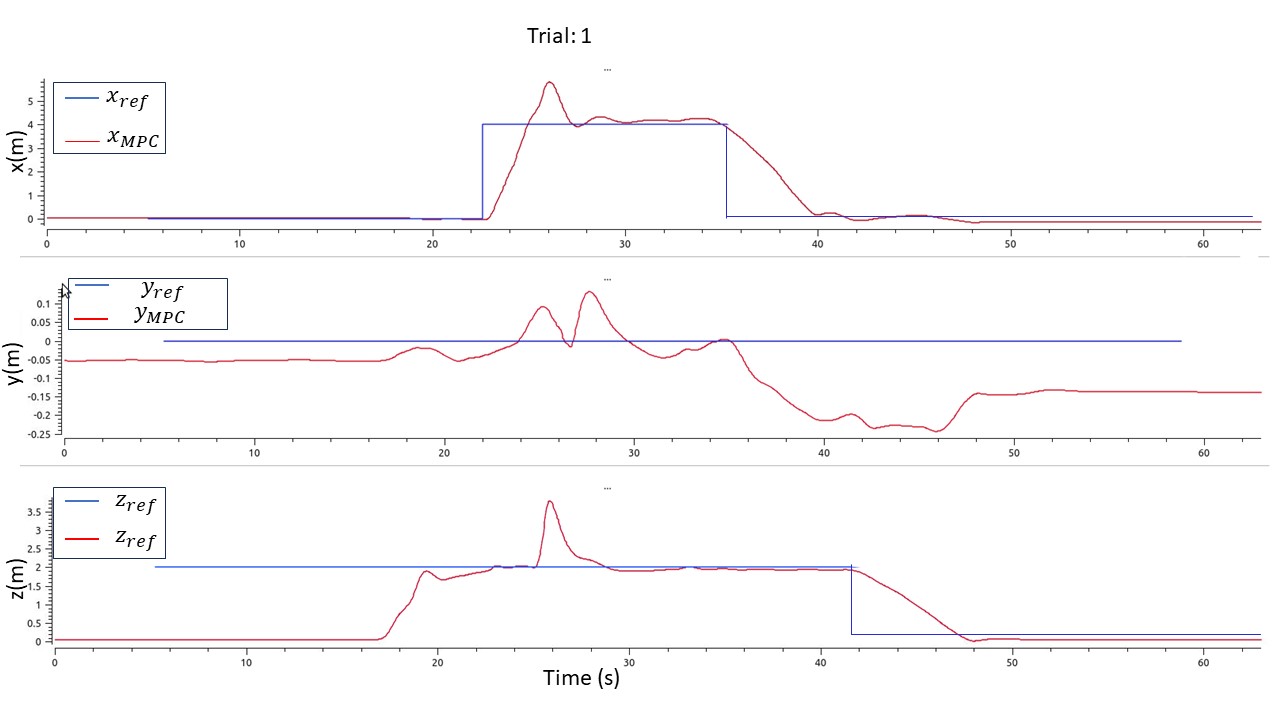}
\centering
\caption{Quadrotor Pose X, Y, and Z Response with MPC Controller}
\label{figure:iteration1pos}
\end{figure}

\begin{figure}[h!]
\centering
\includegraphics[width =8.5 cm ]{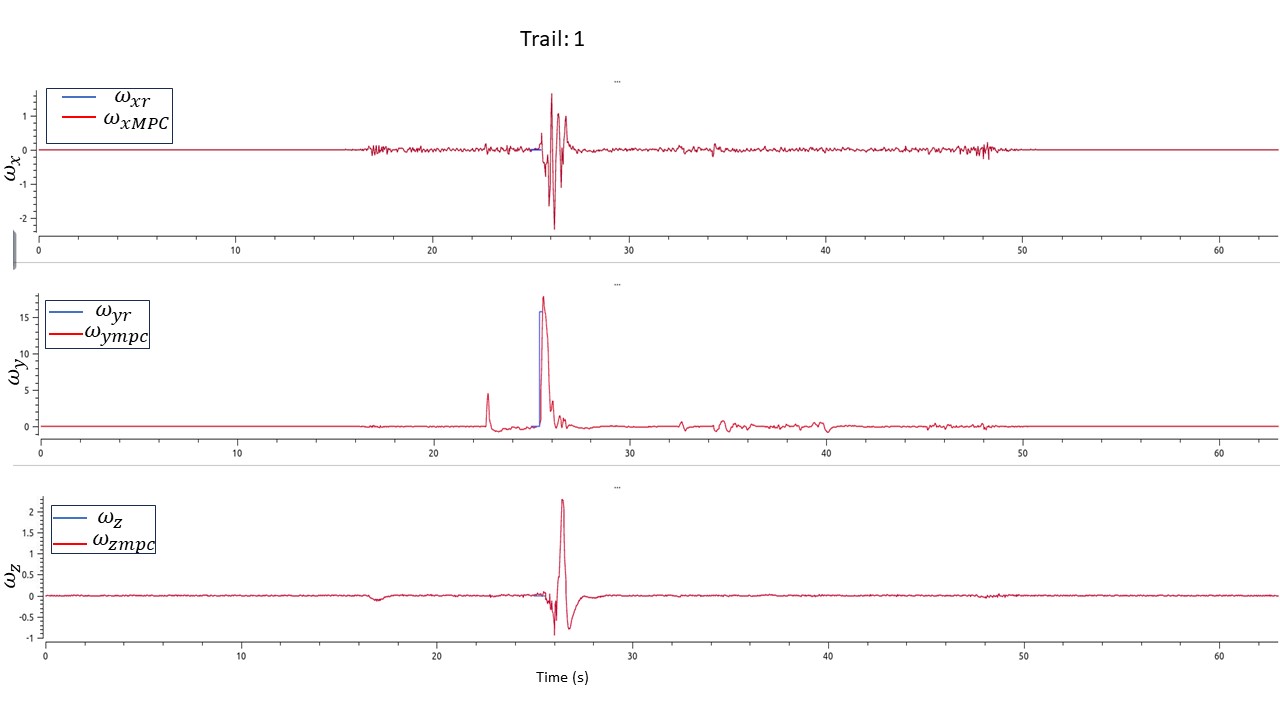}
\centering
\caption{Quadrotor Attitude Rate Response with MPC Controller}
\label{figure:iteration1bodyrate}
\end{figure}

In all trials of the flip maneuver, the mission scenarios and the reference state are expected consistent which is similar to the simulation values. The performance of all the trials are plotted in the graphs and in all the simulation quadrotor is successfully able to execute the flip and able to recover and hover at the desired hold position. One of the trial performances is shown in Fig. \ref{figure:iteration1pos} and \ref{figure:iteration1bodyrate}.

Trial 1: After carefully observing Fig. \ref{figure:iteration1pos}, it can be observed that the flip occurs at 25 seconds, causing a peak in the x and z responses. Following the flip, the MPC controller effectively recovers from the upside-down position and achieves a stable hover of around 30 seconds. The x and z poses of the quadrotor get stabilized before returning to the pre-landing position. Subsequently, the quadrotor successfully lands at the home position.

Let's focus on the attitude rate control during the flip maneuver. The MPC controller successfully tracks the reference of $5\pi$ rad/sec of pitch rate at the beginning of the flip, making an abrupt change within a very short time window. The controller effectively stabilizes the attitude after the flip, demonstrating the effectiveness of the proposed control schemes.

\begin{figure}[h!]
\centering
\includegraphics[width =8.5 cm ]{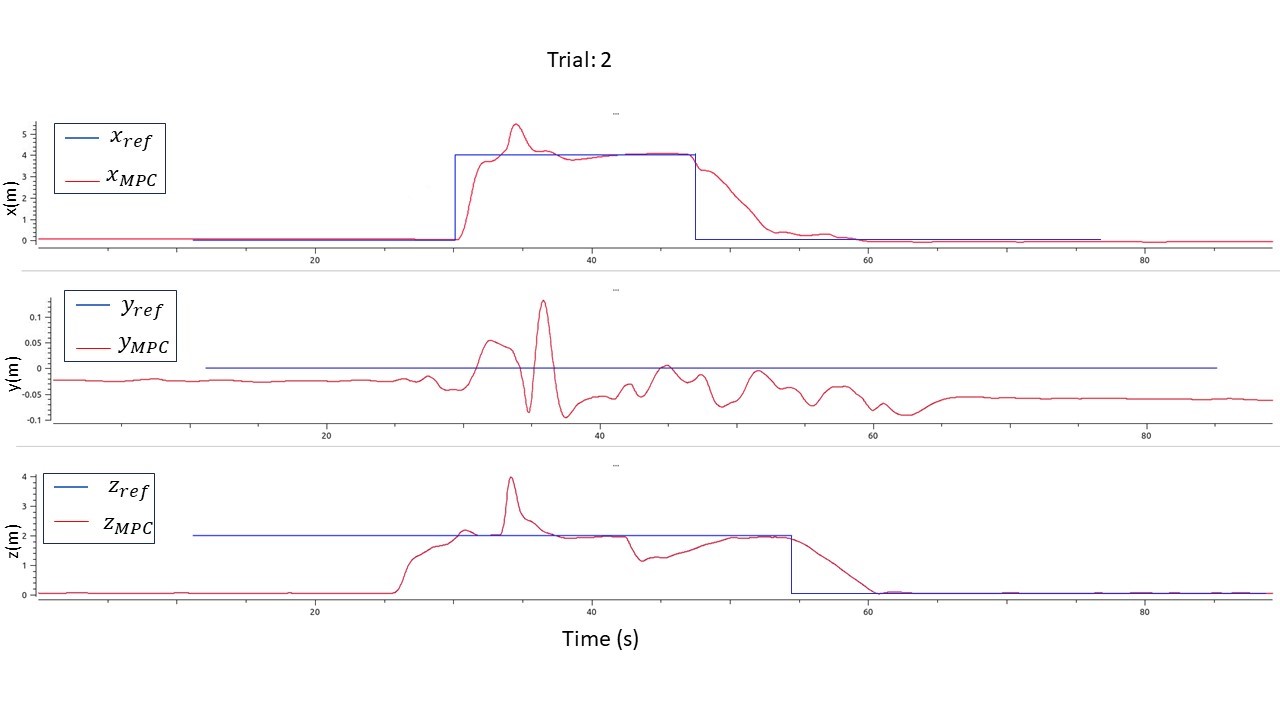}
\centering
\caption{Quadrotor Pose X, Y, and Z Response with MPC Controller}
\label{figure:iteration2pos}
\end{figure}
\begin{figure}[h!]
\centering
\includegraphics[width =8.5 cm ]{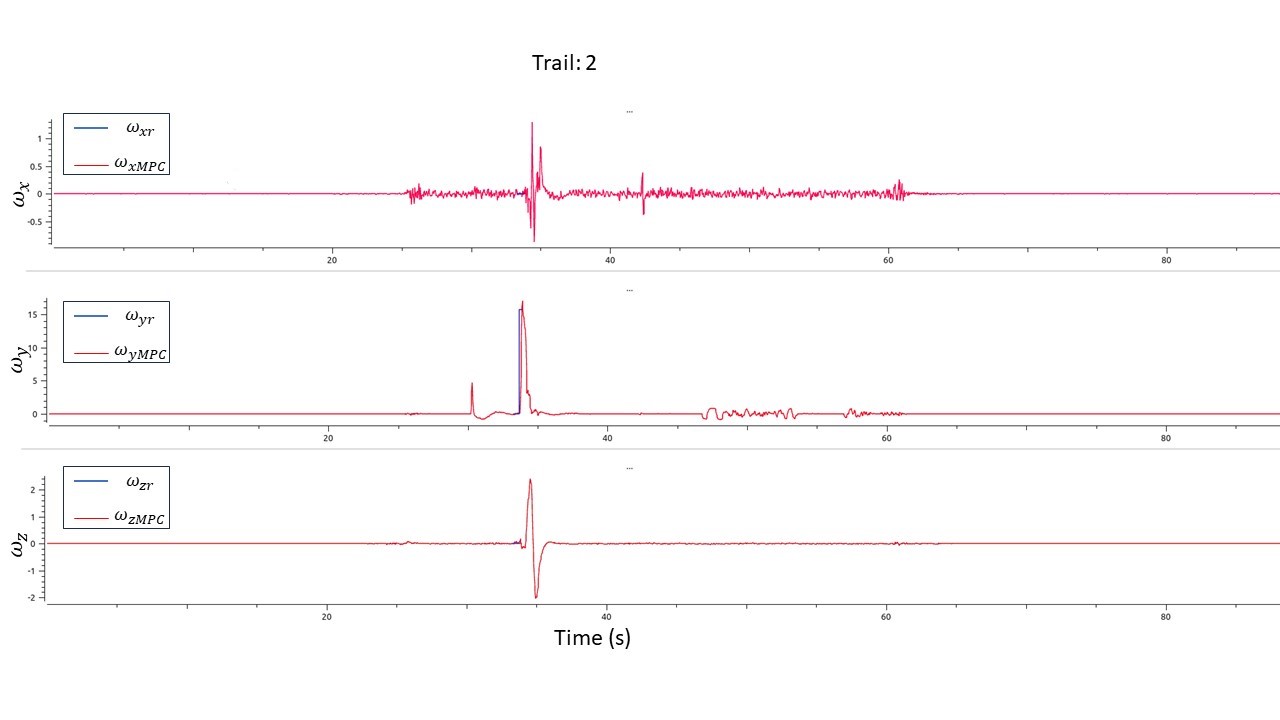}
\centering
\caption{Quadrotor Attitude Rate Response with MPC Controller}
\label{figure:iteration2bodyrate}
\end{figure}
 \begin{figure}[h!]
\centering
\includegraphics[width =8.5 cm ]{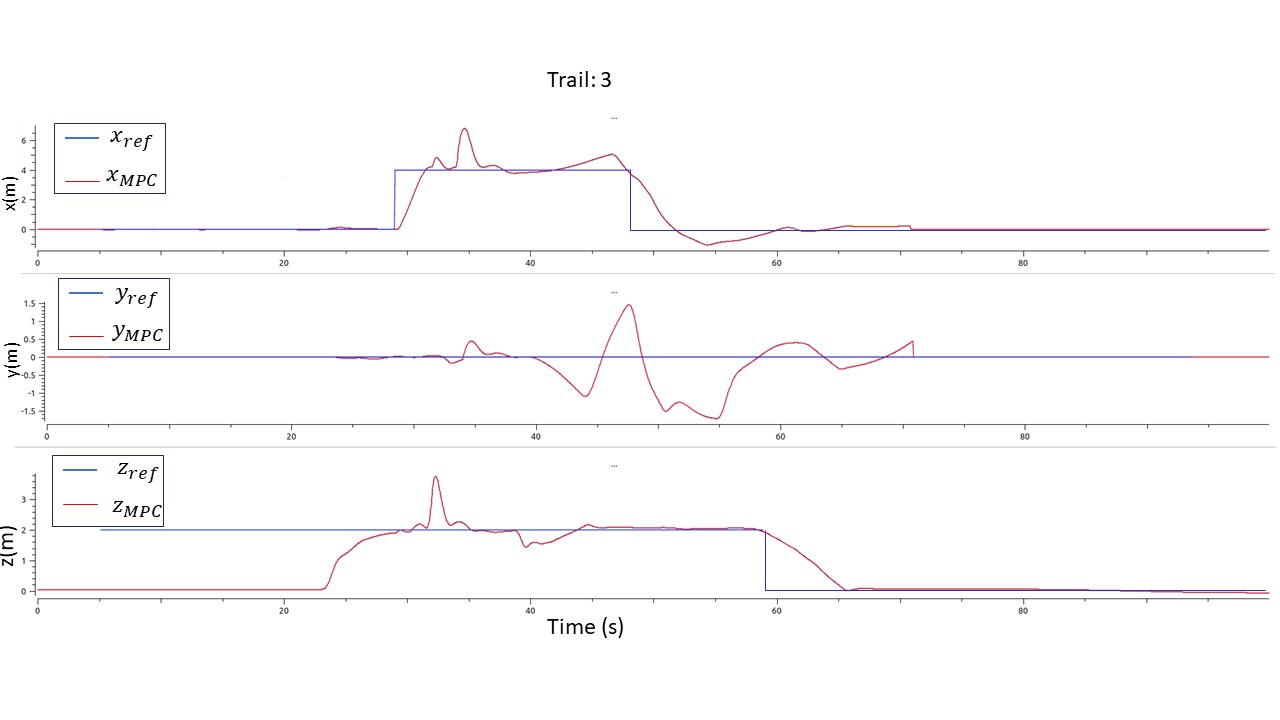}
\centering
\caption{Quadrotor Pose X, Y, and Z Response with MPC Controller}
\label{figure:iteration3pos}
\end{figure}
\begin{figure}[h!]
\centering
\includegraphics[width =8.5 cm ]{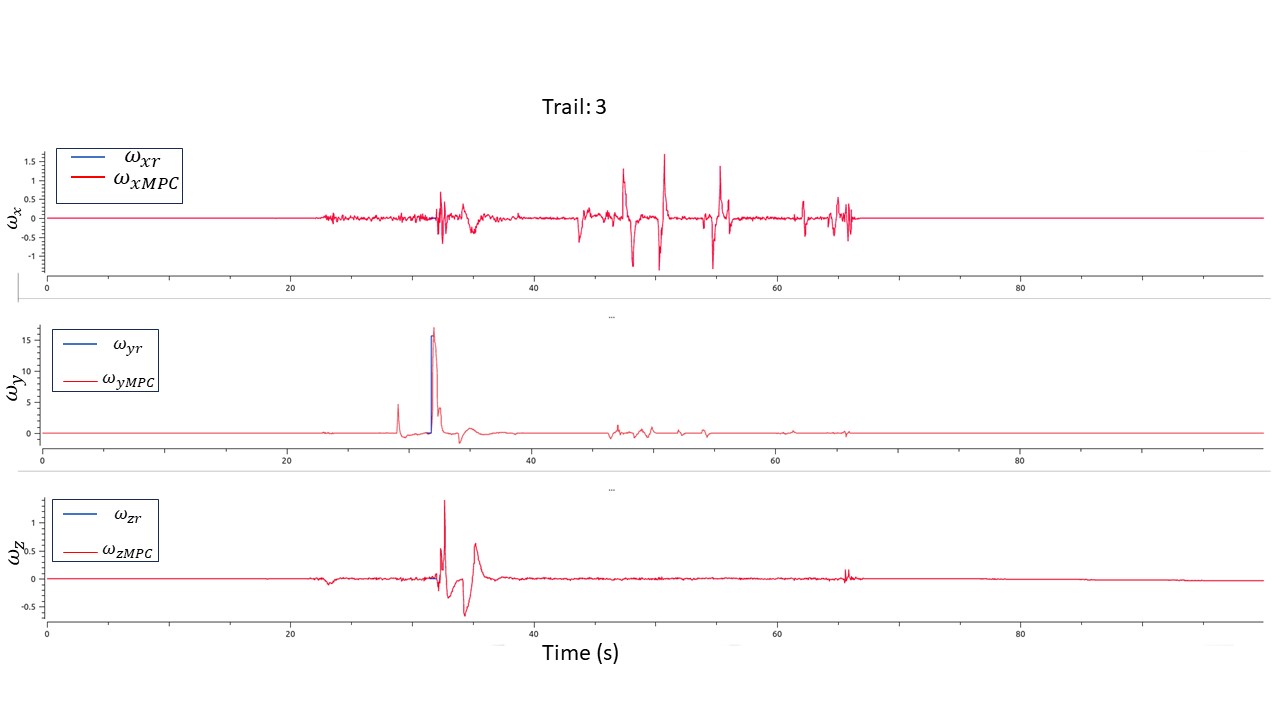}
\centering
\caption{Quadrotor Attitude Rate Response with MPC Controller}
\label{figure:iteration3bodyrate}
\end{figure}
\begin{figure}[h!]
\centering
\includegraphics[width =8.5 cm ]{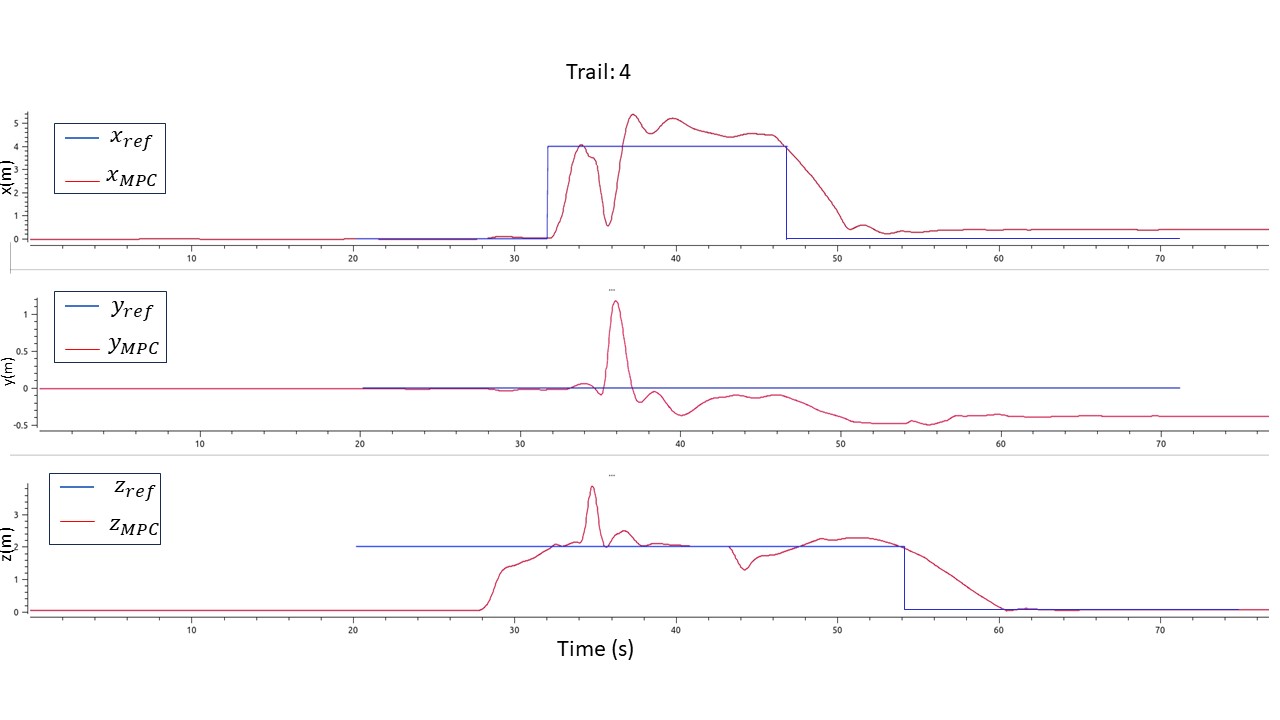}
\centering
\caption{Quadrotor Pose X, Y, and Z Response with MPC Controller}
\label{figure:iteration4pos}
\end{figure}
\begin{figure}[h!]
\centering
\includegraphics[width =8.5 cm ]{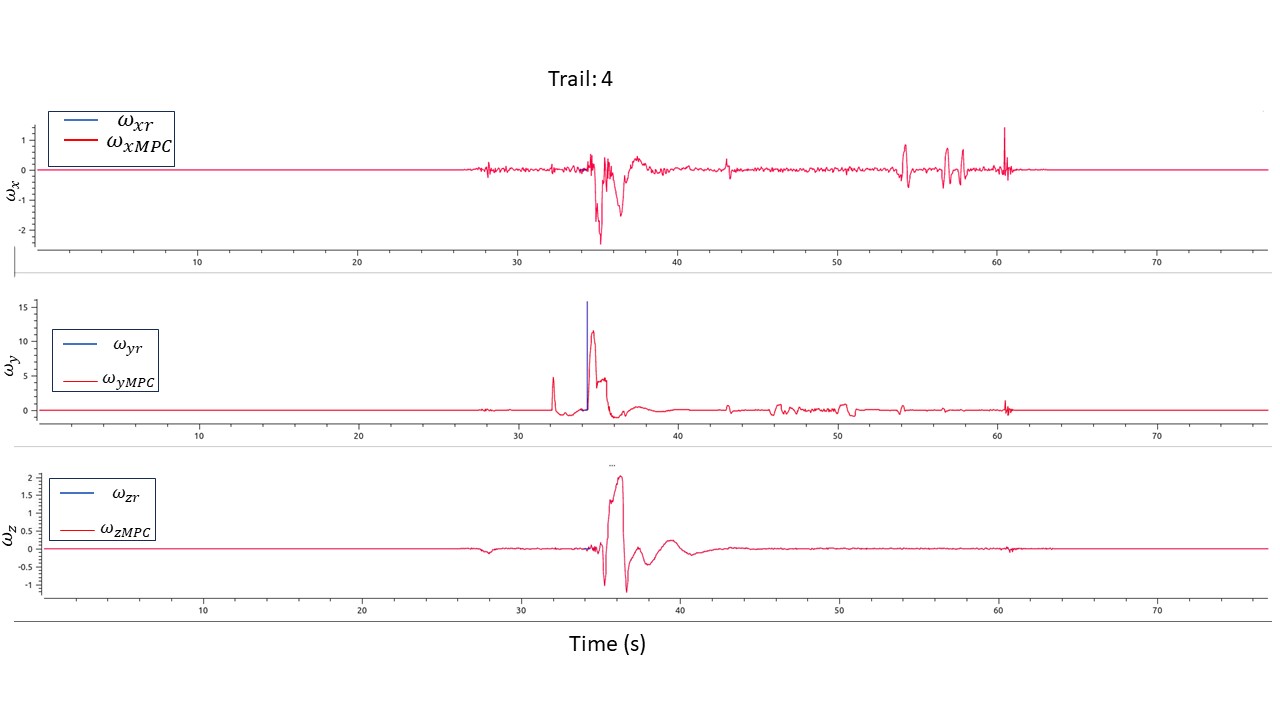}
\centering
\caption{Quadrotor Attitude Rate Response with MPC Controller}
\label{figure:iteration4odyrate}
\end{figure}
\begin{figure}[h!]
\centering
\includegraphics[width =8.5 cm ]{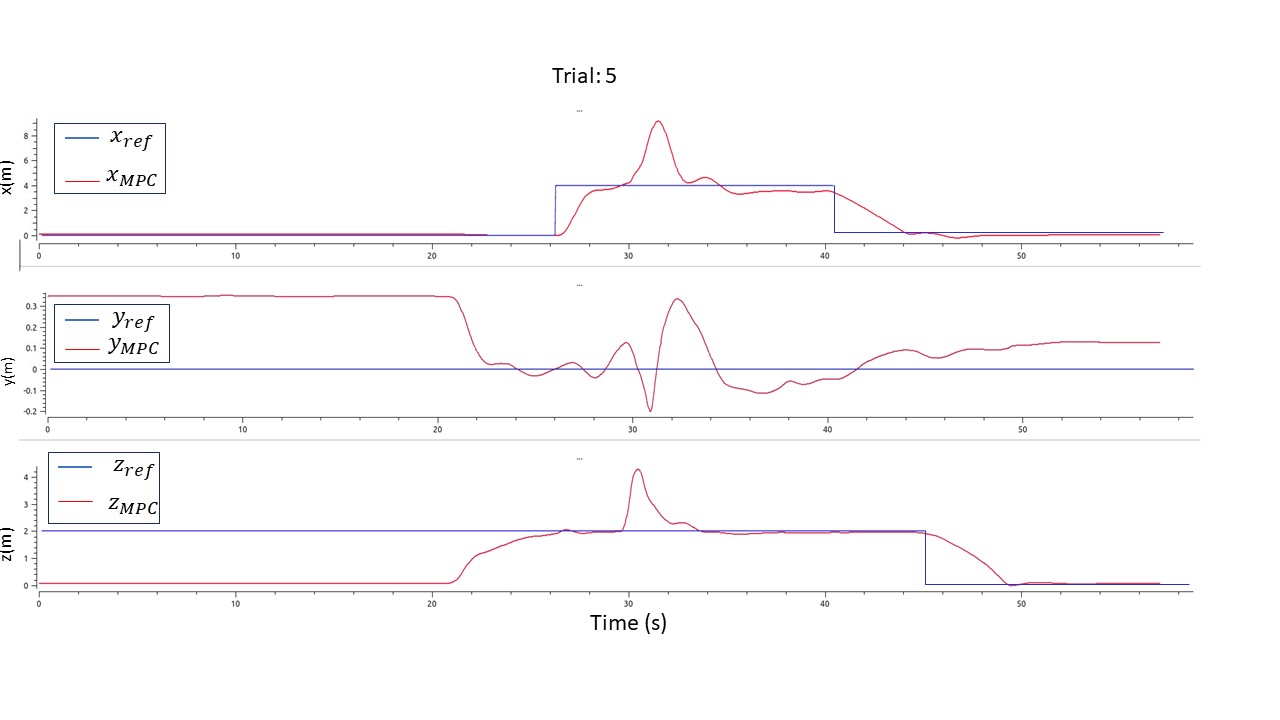}
\centering
\caption{Quadrotor Pose X, Y, and Z Response with MPC Controller}
\label{figure:iteration5pos}
\end{figure}
\begin{figure}[h!]
\centering
\includegraphics[width =8.5 cm ]{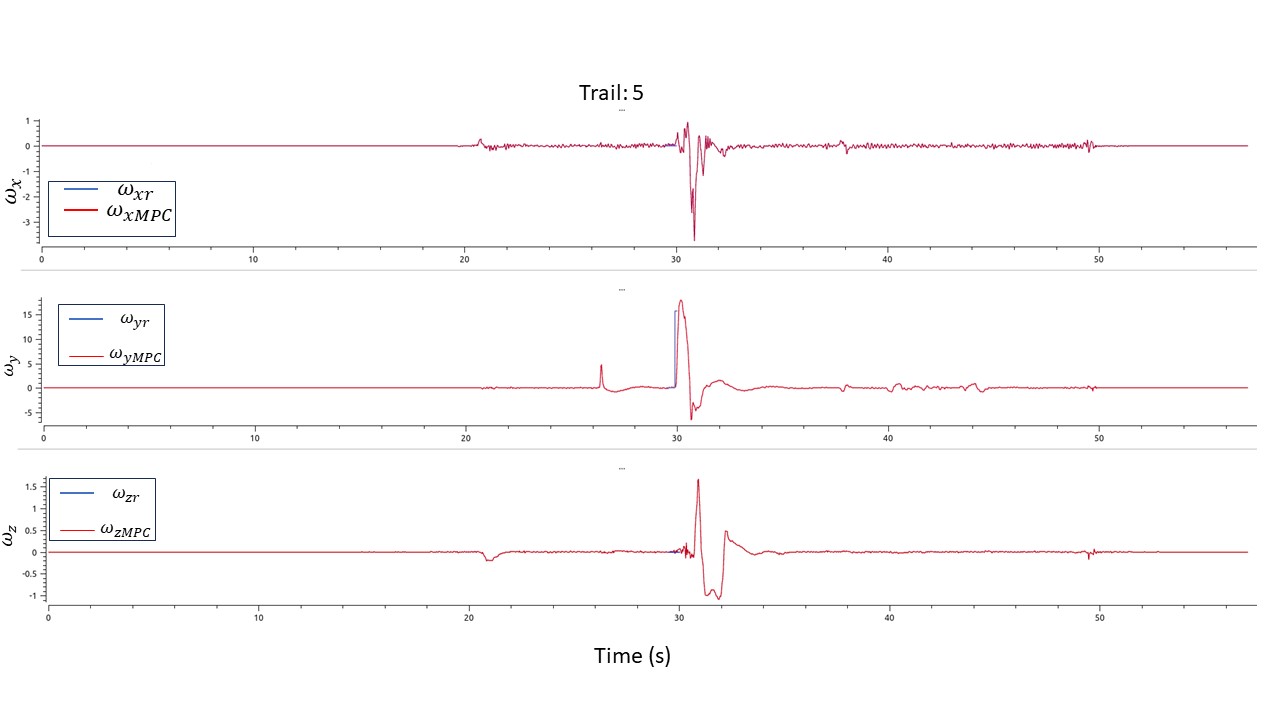}
\centering
\caption{Quadrotor Attitude Rate Response with MPC Controller}
\label{figure:iteration5odyrate}
\end{figure}

\begin{figure}[h!]
\centering
\includegraphics[width =8.5 cm ]{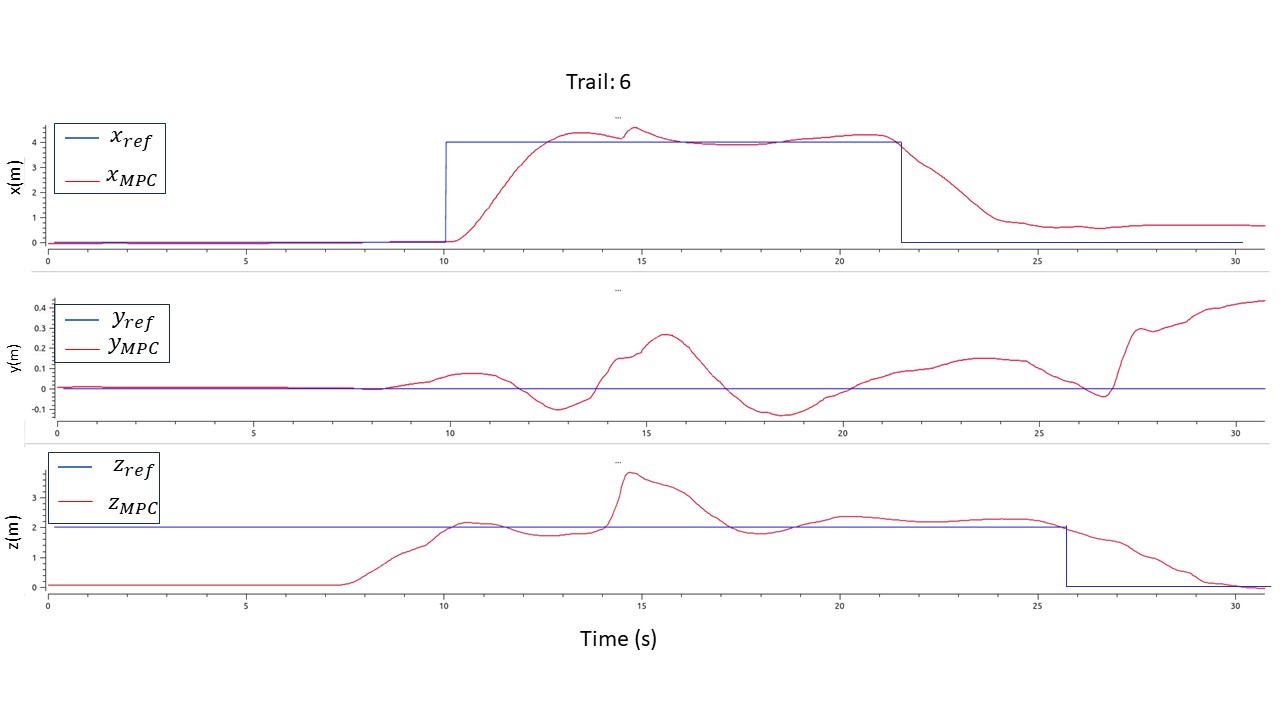}
\centering
\caption{Quadrotor Pose X, Y, and Z Response with MPC Controller}
\label{figure:iteration6pos}
\end{figure}
\begin{figure}[h!]
\centering
\includegraphics[width =8.5 cm ]{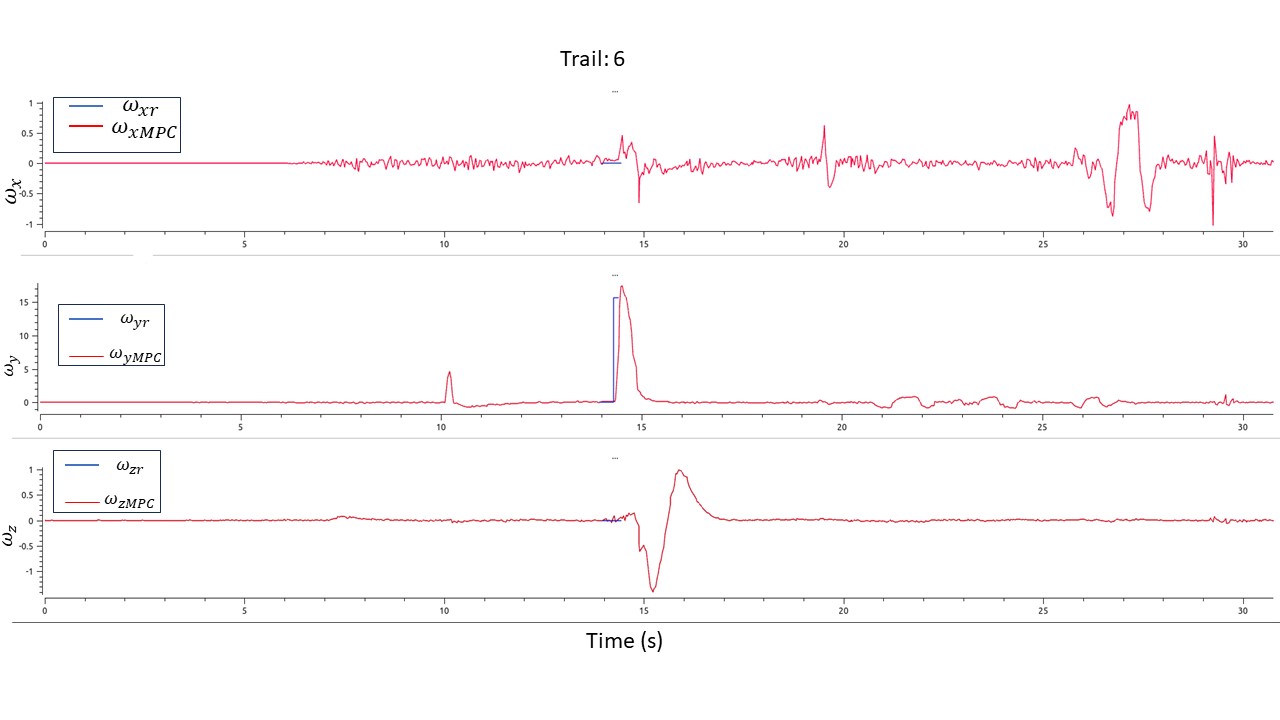}
\centering
\caption{Quadrotor Attitude Rate Response with MPC Controller}
\label{figure:iteration6bodyrate}
\end{figure}

\begin{figure}[h!]
\centering
\includegraphics[width =8.5 cm ]{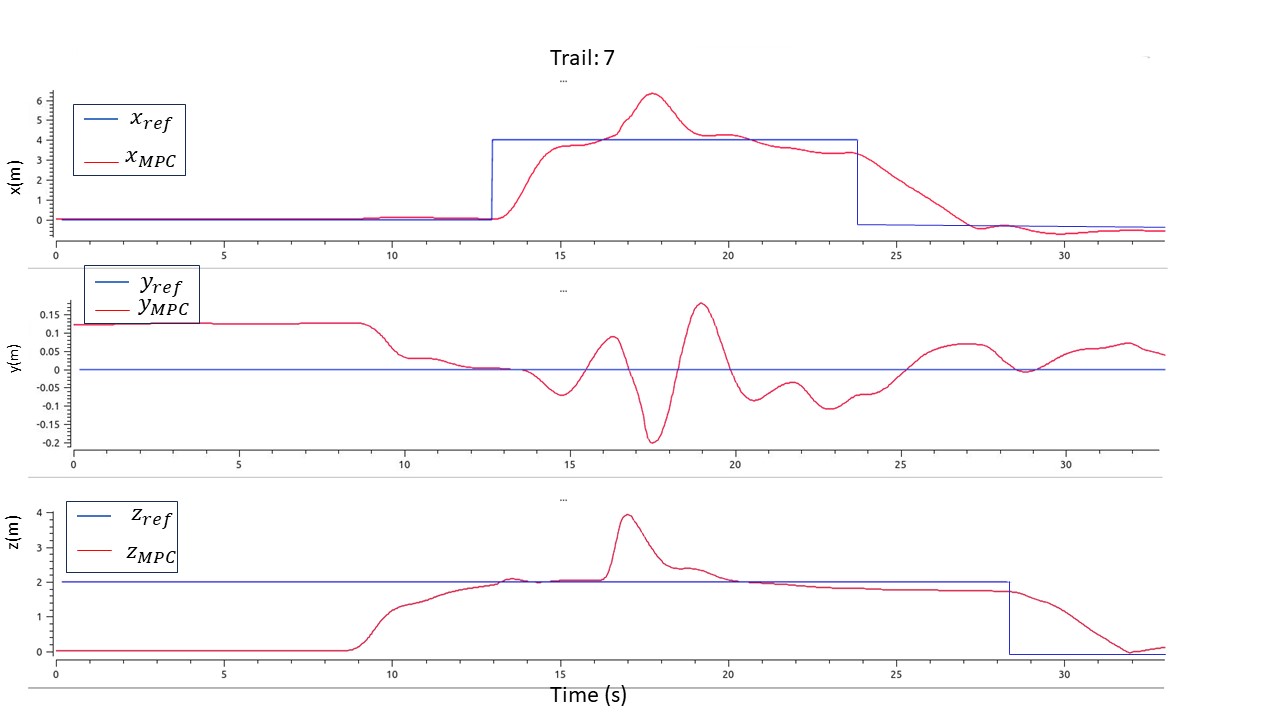}
\centering
\caption{Quadrotor Pose X, Y, and Z Response with MPC Controller}
\label{figure:iteration7pos}
\end{figure}
\begin{figure}[h!]
\centering
\includegraphics[width =8.5 cm ]{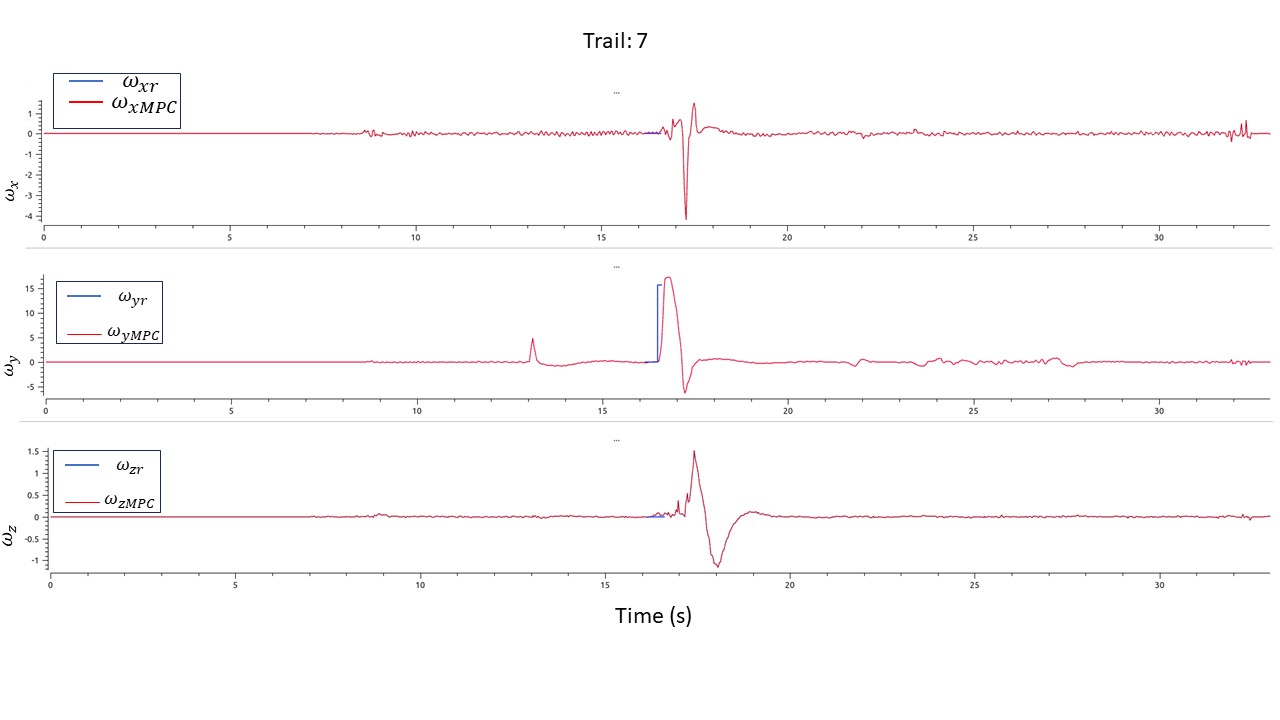}
\centering
\caption{Quadrotor Attitude Rate Response with MPC Controller}
\label{figure:iteration7bodyrate}
\end{figure}

\begin{figure}[h!]
\centering
\includegraphics[width =8.5 cm ]{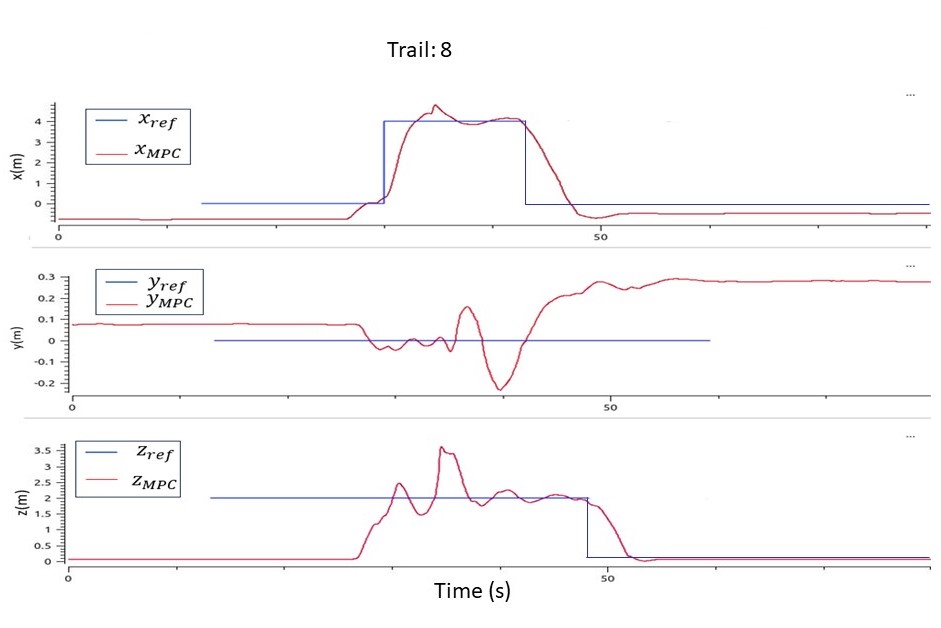}
\centering
\caption{Quadrotor Pose X, Y, and Z Response with MPC Controller}
\label{figure:iteration8pos}
\end{figure}
\begin{figure}[h!]
\centering
\includegraphics[width =8.5 cm ]{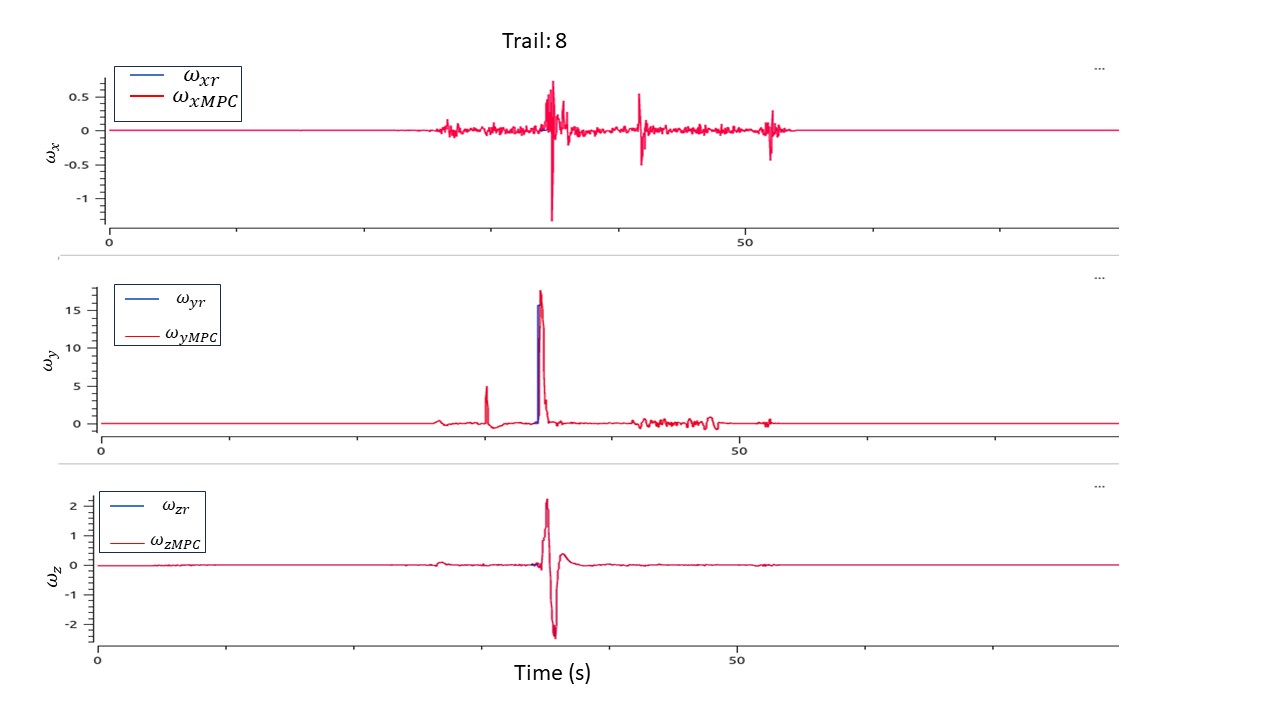}
\centering
\caption{Quadrotor Attitude Rate Response with MPC Controller}
\label{figure:iteration8bodyrate}
\end{figure}

\begin{figure}[h!]
\centering
\includegraphics[width =8.5 cm ]{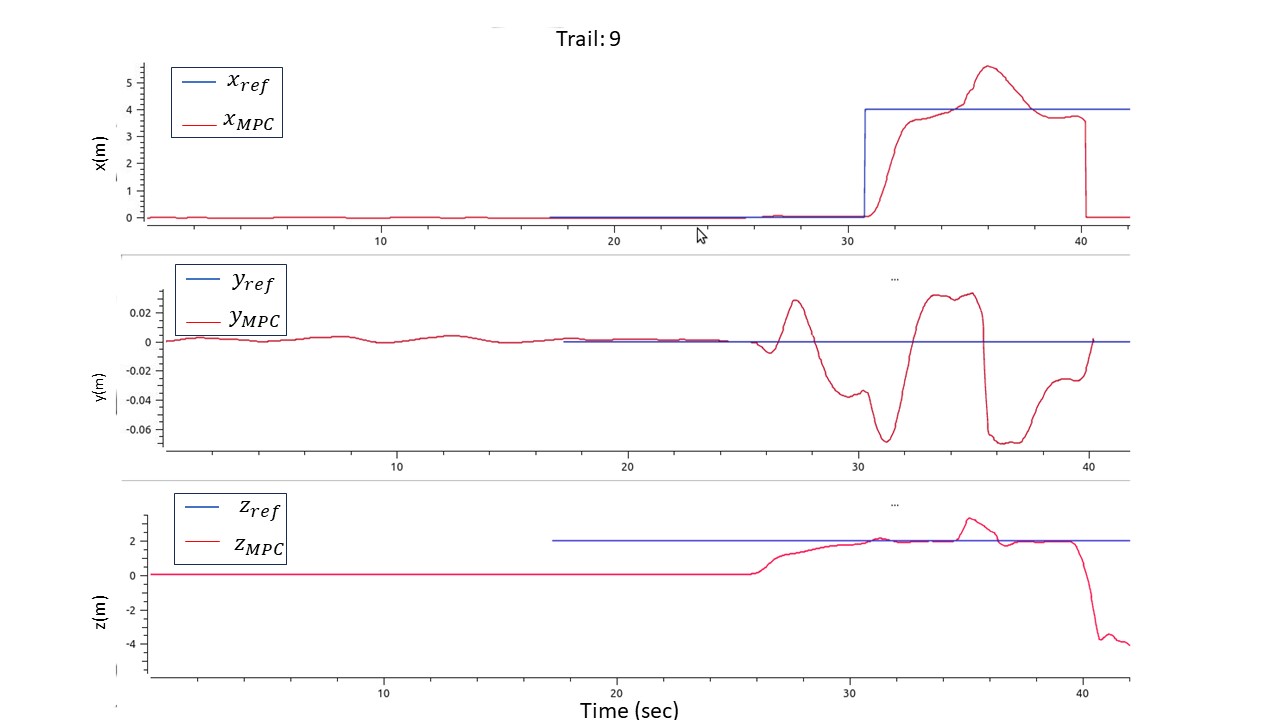}
\centering
\caption{Quadrotor Pose X, Y, and Z Response with MPC Controller}
\label{trialpos}
\end{figure}
\begin{figure}[h!]
\centering
\includegraphics[width =8.5 cm ]{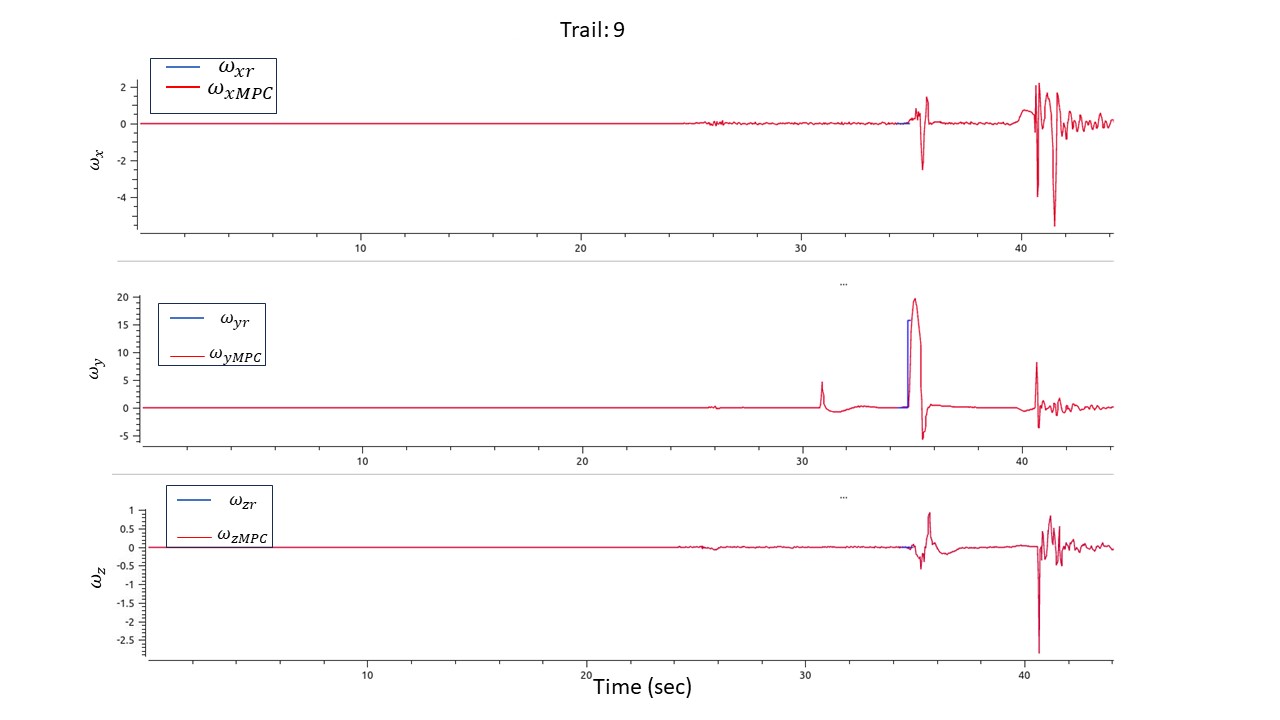}
\centering
\caption{Quadrotor Attitude Rate Response with MPC Controller}
\label{trialbody}
\end{figure}
Trial 2: This trial provides further evidence of the MPC controller's repeatability, as the system's response closely resembles that observed in trial 1. The fact that the system's behavior is highly consistent between these iterations indicates the controller's ability to reproduce desired outcomes consistently. Such repeatability is crucial in various applications, particularly in scenarios where precise and reliable control is necessary. The MPC controller demonstrates its robustness and reliability by achieving similar responses across multiple iterations, reinforcing its effectiveness as a control scheme in the given context.

Trial 3: This trial is another evidence of the repeatability and reliability of the proposed controller owing to the fact that the response of trial 1, 2, and 3 is very similar.

Trial 4: An anomaly was observed during the flip maneuver in this trial. After the flip, the quadrotor exhibited a backward movement from the rally point, as depicted in Fig. \ref{figure:iteration4pos} in the x-direction response. Subsequently, it quickly returned to the hold position. This behavior was attributed to the quadrotor's inability to achieve a sufficient pitch rate to execute the motion similar to Iterations 1, 2, and 3.

As shown in Fig. \ref{figure:iteration4odyrate}, the reference signal for the angular rate (pitch rate) ($\omega_y$) was set to $5/\pi$ rad/s. However, the MPC controller could only achieve 11 rad/s, which proved inadequate to compensate for the dropping speed at the end of the flip maneuver. The lower angular rate resulted in an increased recovery time for the quadrotor, leading to a longer duration for stabilization and regaining control following the completion of the flip.

One factor contributing to this issue was the limitation on gaining altitude. Notably, the flip was initiated at a height of 2 m, while the ASU drone studio, with a ceiling height of 6 m, imposed constraints on altitude gain during MPC design. Consequently, the quadrotor had insufficient time to gain momentum for a faster recovery.

Despite this limitation, the MPC controller demonstrated its ability to stabilize the quadrotor after the flip and execute the mission without a crash. Although the backward movement and longer recovery time were observed, the quadrotor ultimately recovered and maintained control, showcasing the effectiveness of the MPC control scheme within the given constraints.

It is worth noting that this anomaly was observed only in one trial. In the remaining cases, the quadrotor achieved the desired rate and quickly recovered from the flip maneuver.

In the rest of the trials: 5,6,7, and 8, the MPC controller performance is similar to trials 1, 2, and 3. The attitude rate in all the trials reached the reference value and successfully recovered the drone from the flip position. The response of MPC in trials 5,6,7 and 8 is plotted. 

The MPC controller consistently demonstrated successful performance by successfully executing the flip maneuver in all eight iterations. Despite encountering a minor anomaly in trial 4, the quadrotor exhibited exceptional stability and managed to recover without experiencing a crash. This remarkable capability highlights the reliability and resilience of the MPC controller in handling challenging and aggressive maneuvers, such as flips.

The series of iterations not only serves as a testament to the effectiveness of the MPC controller but also establishes a strong foundation for the execution of the flip and throw maneuver. The controller's ability to navigate through various iterations with consistent success showcases its robustness in the face of unexpected disturbances and uncertainties.

These findings provide valuable insights into the MPC controller's reliability, affirming its suitability for advanced applications that involve precise control and dynamic movements. The successful execution of the flip maneuver in multiple iterations bolsters confidence in the controller's performance and further strengthens its credibility for future mission-critical tasks.
Overall, the consistent and reliable performance of the MPC controller in executing the flip maneuver, even in the presence of occasional anomalies, underscores its robustness and establishes it as a promising control scheme for accomplishing complex aerial maneuvers with precision and safety.

Experiments conducted in real-time serve as invaluable learning experiences, and play a crucial role in the debugging and enhancement of quadrotor flights and the implemented algorithms. In the context of trial 9, as shown in Fig. \ref{trialpos} and Fig. \ref{trialbody}, where testing occurred within the MOCAP environment, the quadrotor effectively executed the assigned mission scenario. However, an issue arose during the flip recovery phase, as the quadrotor lost the estimation provided by the MOCAP system. Consequently, the fail-safe mechanism was triggered, causing the quadrotor flight controller to switch to 'return' mode.

While operating in 'return' mode, the quadrotor autonomously ascended to a predetermined height and safely landed at its designated home position. The estimation loss occurred due to the absence of a marker on the bottom of the quadrotor. This hindered the MOCAP system's ability to accurately localize the quadrotor when it was in an upside-down position. As a result, the MOCAP system failed to track the quadrotor's position, leading to the estimation loss.

A solution was implemented to rectify this drawback and ensure accurate estimation during flips. Markers were strategically added to the sides of the quadrotor arm and the bottom surface. These additional markers enabled the MOCAP system to maintain localization even when the quadrotor was inverted. The quadrotor's ability to execute flips and maintain reliable estimation throughout the flight was significantly improved by addressing this limitation.

\section{Conclusion}
This work, comprehensively explores the flip maneuver, covering its dynamics, control design, real-time implementation, and its extension to the more complex flip and throw maneuver. Integrating the probe and quadrotor dynamics proves crucial in designing control strategies for aggressive maneuvers like the flip. The performance of the MPC controller is thoroughly tested using ROS and gazebo integration in simulation. The CasADi toolbox in Python is employed to solve the optimization problem, utilizing the Sequential Quadratic Programming (SQP) solver method. The MPC controller successfully reduces trajectory tracking error to less than 10 $\%$ 
Furthermore, the control scheme is implemented in real-time using the MOCAP system for localization and a companion computer for computation. The robustness and repeatability of the MPC controller are verified by performing eight iterations of the flip maneuver in real-time, demonstrating its ability to execute all iterations successfully without crashes and achieve a tracking error reduction of less than 10$\%$.

Building upon understanding the flip maneuver.The MPC controller successfully executes the flip and throw maneuver, accurately landing the probe on the desired target coordinate. This comprehensive analysis showcases the robustness and reliability of the proposed control design, demonstrating the effectiveness of the MPC controller in both simulation and real-time scenarios. The findings of this work contribute to the advancement of control strategies for complex aerial maneuvers, providing valuable insights into the capabilities and potential applications of MPC in various domains.


%



%

\bibliographystyle{plain} 
\bibliography{lib} 

%




\end{document}